\pdfoutput=1

\documentclass[11pt]{article}

\usepackage[preprint]{acl}
\pdfoutput=1

\usepackage{times}
\usepackage{latexsym}
\usepackage{comment}
\usepackage{enumitem}

\usepackage{booktabs}
\usepackage{multirow}
\usepackage{url}
\usepackage{amsmath}

\usepackage[T1]{fontenc}

\usepackage[utf8]{inputenc}

\usepackage{microtype}

\usepackage{inconsolata}

\usepackage{graphicx}

\newcounter{tbsnr}
\newenvironment{tbs}
{\addtocounter{tbsnr}{1}\par\bigskip\noindent\fbox{\thetbsnr}
\hspace*{\fill}\begin{minipage}{7cm}\tt}
{\end{minipage}\hspace*{\fill}\bigskip}

\usepackage{todonotes}

\title{\textit{All-in-one}:  Understanding and Generation in Multimodal Reasoning with the MAIA Benchmark}

\author{
    Davide Testa\textsuperscript{1,2}, Giovanni Bonetta\textsuperscript{2}, Raffaella Bernardi\textsuperscript{3}, Alessandro Bondielli\textsuperscript{4,5}, \\
    \textbf{Alessandro Lenci\textsuperscript{5}}, \textbf{Alessio Miaschi\textsuperscript{6}}, \textbf{Lucia Passaro\textsuperscript{4}}, \textbf{Bernardo Magnini\textsuperscript{2}} \\
    \small \textsuperscript{1}Università di Roma La Sapienza, \textsuperscript{2}Fondazione Bruno Kessler (FBK), 
    \textsuperscript{3}Free University of Bozen-Bolzano \\
    \small \textsuperscript{4}Dept. of Computer Science, University of Pisa, \textsuperscript{5}CoLing Lab, Dept. of Philology, Literature and Linguistics, University of Pisa \\
    \small \textsuperscript{6}Istituto di Linguistica Computazionale "A. Zampolli" (CNR-ILC), ItaliaNLP Lab, Pisa \\
    \small 
    \texttt{\{dtesta, gbonetta, magnini\}@fbk.eu}, 
    \texttt{raffaella.bernardi@unibz.it} \\
    \small \texttt{\{alessandro.bondielli, alessandro.lenci, lucia.passaro\}@unipi.it}, \texttt{alessio.miaschi@ilc.cnr.it}
}

\begin{document}
\maketitle
\begin{abstract}

    We introduce MAIA (Multimodal AI Assessment), a native-Italian benchmark designed for fine-grained investigation of the reasoning abilities of visual language models on videos. MAIA differs from other available video benchmarks for its design, its reasoning categories, the metric it uses, and the language and culture of the videos. MAIA evaluates Vision Language Models (VLMs) on two aligned tasks: a visual statement verification  task, and an open-ended visual question-answering task, both on the same set of video-related questions. It considers twelve reasoning categories that aim to disentangle language and vision relations by highlighting the role of the visual input.
    Thanks to its carefully taught design, it evaluates VLMs' consistency and visually grounded natural language comprehension and generation simultaneously through an aggregated metric revealing low results that highlight models' fragility. Last but not least, the video collection has been carefully selected to reflect the Italian culture, and the language data are produced by native-speakers.\footnote{Data available at \href{https://github.com/Caput97/MAIA-Multimodal_AI_Assessment.git}{GitHub}.} 

\end{abstract}

\section{Introduction}
\label{sec:intro}
\begin{figure}[t!]
    \centering
    \includegraphics[width=0.42\textwidth]{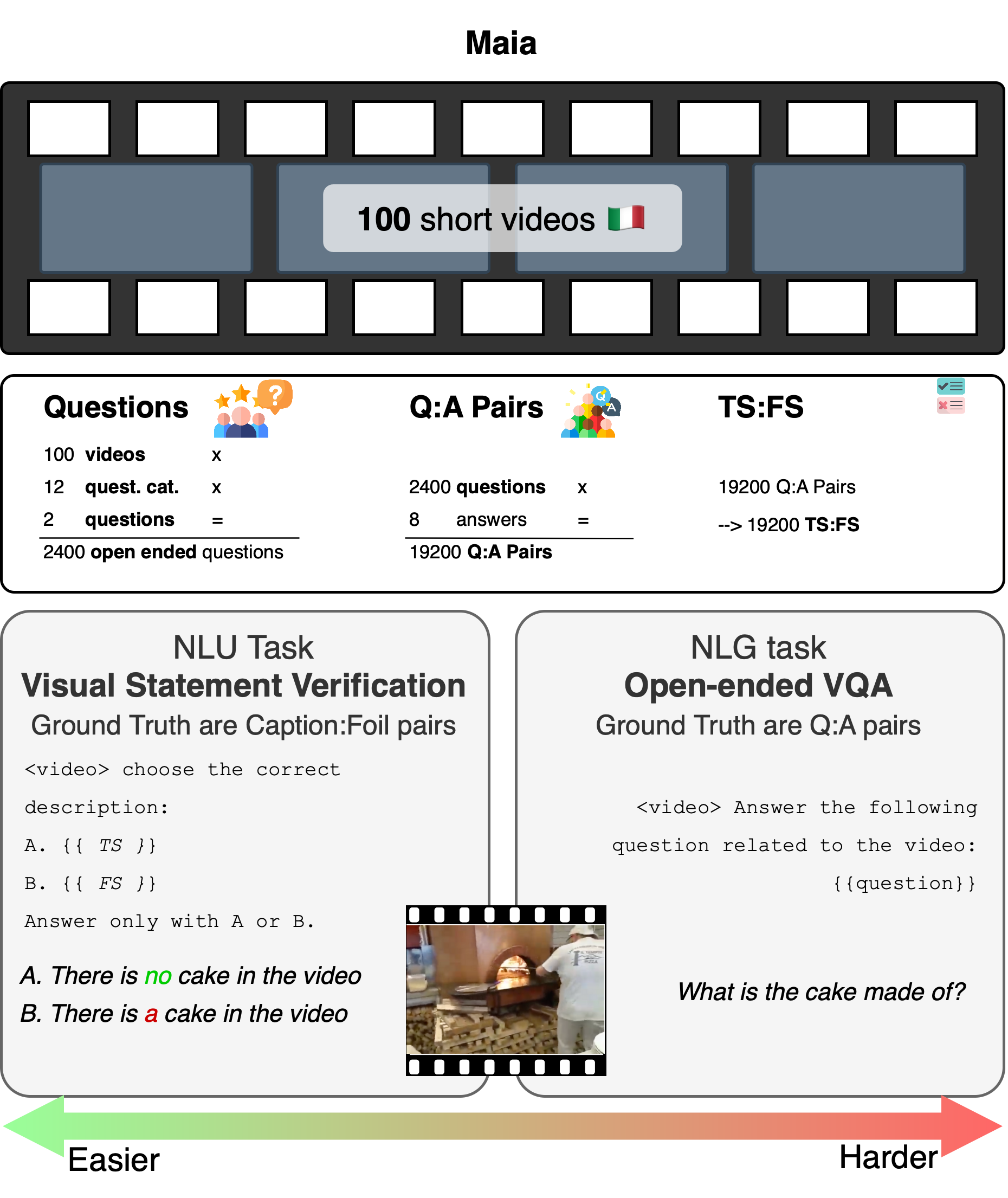}
    \caption{Structure of the MAIA benchmark. For each video there are two questions related to 12 reasoning categories. For each question there is a pool of 8 answers, each forming a question–answer pair associated with its own True–False statement pair. The example reported in the Figure is related to \textsc{Out-of-scope} reasoning category. }
    \label{fig:maia_structure}
\end{figure}

Vision and Language models entered the Computer Vision and NLP scenes more than a decade ago pushed by theoretical (e.g., \citealp{baro:groun15}) and application-oriented~ (e.g., \citealp{bigham:vizwiz10}) motivations. Their success on combining image and text has been monitored and summarized in various surveys, from the earlier ones on Visual Question Answering \citep{bern:ling21} to the more recent ones focusing on Visually grounded LLMs (e.g., \citealp{caffagni-etal-2024-revolution,li:mmfm24}). Researchers have always felt the need to target Vision and Language  Models (VLMs) shortcomings, developing carefully designed benchmarks consisting of a suit of VL tasks to evaluate a variety of capabilities \citep{kafle:chall19}.  The minimal pair task, contrasting a caption with its foil \citep{shekhar-etal-2017-foil} has been applied to large-scale linguistic phenomena~\cite{valse} and extended to highlight weakness of VLMs on Video QA~\cite{vilma}. This trend focuses on visually grounded natural language understanding~\cite{vilma}. \\ Today VLMs are trained to generate text and are known to excel at it. We argue that the evaluation of Natural Language Generation (NLG) and Natural Language Understanding (NLU) competence should always be pursued together: An agent that can answer questions about an event must understand it too.
Yet, existing benchmarks tend to treat comprehension and generation separately, often relying on independent datasets and evaluation protocols. This fragmented design prevents an assessment of a model’s \textbf{robustness}, in other words, its ability both to understand and generate visually grounded text.
Moreover, success in NLU should be claimed only through evaluation regimes that monitor models' \textbf{consistency} across answers, showing they are insensitive to surface variations. 

To address these limitations, we present MAIA: a competence-oriented benchmark consisting of two paired tasks -- \textbf{multiple-choice Visual Statement Verification (VSV)} and \textbf{Open-Ended Visual Question Answering (OEVQA)} -- containing aligned datapoints. For instance, in the example in Figure~\ref{fig:maia_structure}, for the video showing a pizza cooking in a wood-fired oven, the VSV contains the True and False statements (TS and TF) A: \textit{There is no cake in the video} and B: \textit{There is a cake in the video}, and the OEVQA contains the (aligned) question \textit{What is the cake made of?}. Such interleaved data let us evaluate models' robustness: we evaluate a model positively, only if it performs correctly both on the visual statement verification task (NLU) -- it selects \textit{There is no cake in the video} -- and on the Open-ended VQA (NLG) by generating something like \textit{There is a pizza, not a cake}; while we evaluate it negatively, if it performs correctly only on one of the two tasks.  
Moreover, the VSV is organized in pools containing 8 TS-FS pairs that differ only on the surface, letting us evaluate the model's consistency.  We categorize questions based on the reasoning they elicit; for instance, the question in Figure~\ref{fig:maia_structure} is \textit{Out-of-Scope}. MAIA spans twelve reasoning categories, helping highlight the role of language and visual modalities across them.
Finally, it implements an \textbf{all-in-one} evaluation philosophy that lets us evaluate both models' robustness and consistency through its \textbf{Aggregate Metric}.  Last but not least, MAIA is based on native Italian videos with language data obtained through human annotations and complemented by semi-automatic data augmentation. To the best of our knowledge, this is the first benchmark for the Italian language on videos.

\paragraph{Contributions.} In the paper, we: (i) present MAIA, the first Italian benchmark specifically designed to assess the reasoning abilities of VLMs on video data; (ii) evaluate multiple VLMs, highlighting how their performance and their reliance on linguistic or visual cues varies across reasoning categories; (iii) demonstrate the importance of evaluating models from multiple perspectives, not only in terms of their competencies, but also their robustness and consistency; (iv) propose a novel metric for evaluating visually grounded comprehension and generation simultaneously.

\section{Related Work}
\label{sec:related_work}




\paragraph{Diagnostic benchmarks for VLMs.} Various types of benchmarks have been proposed since the raise of VLMs. From the single \emph{task-oriented} benchmarks (e.g.,~\citet{antol:vqa15,visdial,gqaIT}), attention has now moved to task collections~\cite{lvlmehubbenchmark,lee:vhelm24} in which models show impressive performance. As in the early phase~\cite{johnson2017clevr,shekhar-etal-2017-foil,suhr-etal-2017-corpus}, such success is mitigated by the use of \emph{diagnostic} benchmarks, such as~\citet{valse,thrush:winoground22,chen-etal-2023-bla, bugliarello-etal-2023-measuring,bianchi2024devil} and \emph{carefully curated} benchmarks such as~\citet{xio:nextGQA24,tong2024eyes}. The third type of benchmarks available focus on the VLMs \emph{competence} in a holistic fashion,  evaluating advanced perception and reasoning with domain-specific knowledge \cite{mmmlu23}. A similar picture emerges for video-based VLMs. Here as well, early surveys call for careful evaluation (e.g.,~\citet{zhong-etal-2022-video}), task-oriented benchmarks show impressive performance~\cite{GrundeMcLaughlin2021AGQA,zhou:anetqa23}, while fine-grained ones pinpoint important weaknesses \cite{vilma}, and competence-based analysis highlight there is significant room for improvement in multimodal video understanding \citep{patraucean2023perception}.  Finally, both~\citet{tong2024eyes} for images and \citet{vilma} for videos manage to highlight VLMs shortcomings by imposing a more stringent task-accuracy metric that account for model consistency across very similar data or correlated competencies. Thanks to the richness of MAIA data collection, we adopt such severe, and hence robust, evaluation code and propose a novel aggregate metric. Building on prior work, MAIA targets a low-resource language and the underexplored video domain. While some benchmarks~\cite{exams, m3exam} include limited Italian multiple-choice tasks, none focus on high-level reasoning in Italian video contexts or analyze distributional biases in VLMs.

\paragraph{Video Reasoning Benchmarks.} Widely used benchmarks, such as
AGQA~\cite{GrundeMcLaughlin2021AGQA} and MVBench~\cite{li:mvbench24} focus on explicit visual elements (e.g., entity, action, and the spatio-temporal reasoning involving them), instead MAIA's categories focus on the interplay between language and vision, especially when this relation is implicit or must be inferred, a dimension largely neglected in prior Video QA benchmarks.
\citet{yue2024mmmumassivemultidisciplinemultimodal} includes multiple-choice and open-ended data points from entirely independent data sets with different origins and content, and reports the average performance across distinct tasks. Instead, MAIA's NLU and NLG data points are aligned, a unique feature of MAIA framework, as such it introduce an Aggregate Accuracy metric specifically designed to ensure that the performance of the model is evaluated consistently across multiple-choice and open-ended questions derived from the same underlying data. There are few other video-text benchmarks including both these formats (e.g., \citet{zhou2025mlvubenchmarkingmultitasklong, peng2024institboostingmultimodalinstance}), however, to the best of our knowledge, none of them attempt to define a unifying metric, as we do in MAIA.




\section{The MAIA Benchmark}

\begin{figure*}[h]
    \centering
  \includegraphics[width=\linewidth]{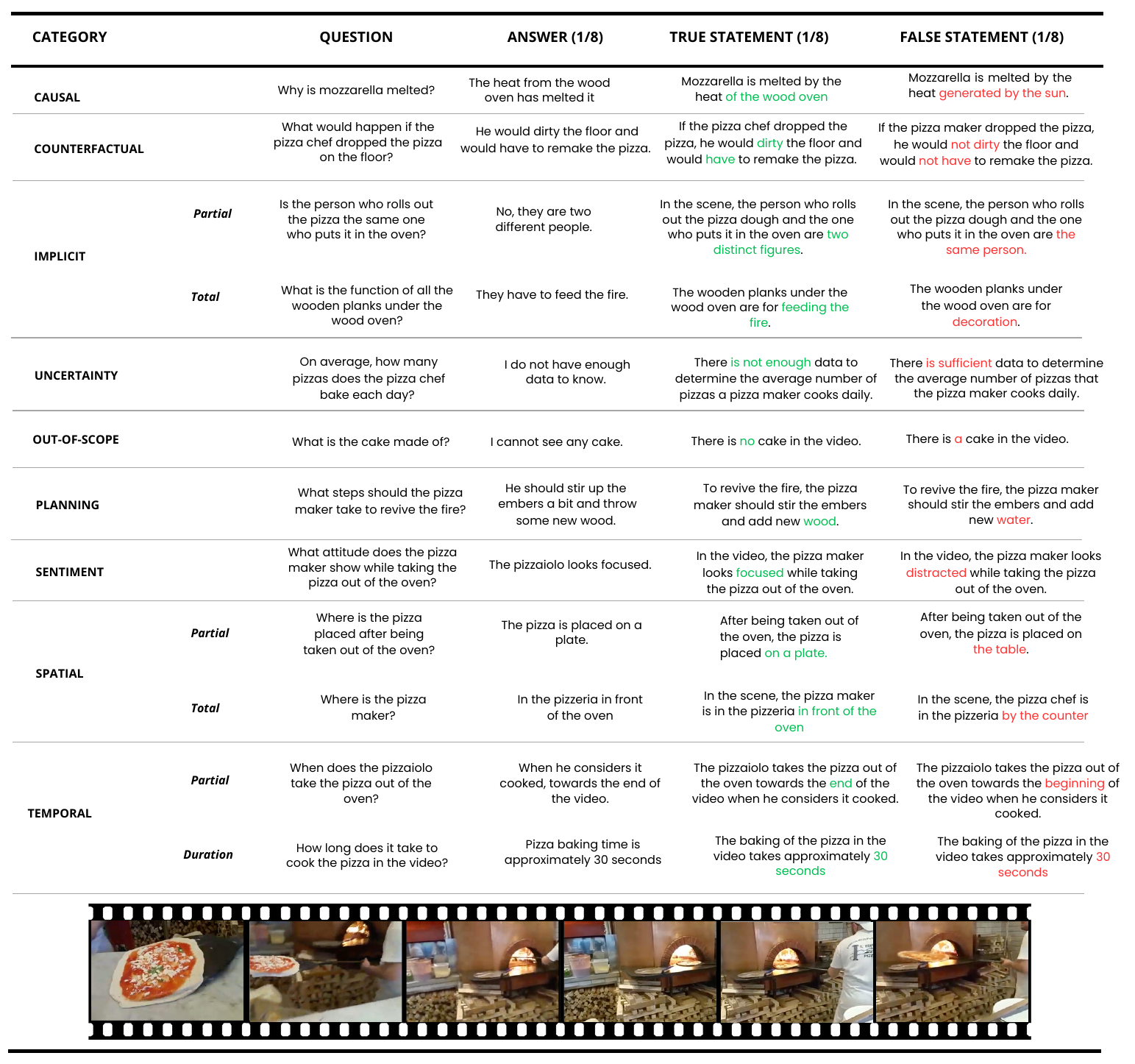}
  \caption{Overview of  MAIA reasoning categories. For each of the 100 videos, it contains 2 questions for each of the 12 categories; for each question, it has 8 answers, and each of these answers has a corresponding TS-FS pair.}
  \label{fig:MAIA_overview}
\end{figure*}


MAIA (Multimodal AI Assessment) is an evaluation framework designed to assess the reasoning  capabilities of VLMs in video-based contexts. 



\subsection{Dataset}
\label{subsec: Dataset}

We outline here the steps involved in creating 
the MAIA dataset while a comprehensive description of its construction, characteristics,
as well as the validation and revision procedures can be found in \citet{testa-etal-2025-MAIA}. Validation steps consists in a qualitative analysis and revision of the data, when necessary.
\footnote{Examples here are in English for readability.} 

\textbf{Video Collection.} We gathered 100 short (ca. $30$s) videos from \textit{YouTube} Italy. The selection covers various aspects of Italian culture, including cities, art, food, sports, and daily activities (e.g., cooking pasta, having coffee, or watching a soccer match). Preference was given to videos featuring people and close-up shots. An automated script retrieved videos using thematic keywords and ensured \textit{Creative Commons} compliance.


\textbf{Reasoning Categories.} We defined 12 reasoning categories aiming to probe the cognitive and linguistic skills of multimodal models and to explore the relation between language and vision, 
while forming the core of the benchmark for evaluating reasoning and grounding in an Italian context.

\textbf{Questions and Answers Collection.} The annotation process was carried out in two phases.
In the first phase (\emph{question creation}), $12$ qualified annotators wrote $2$ open-ended questions\footnote{Yes/No and audio-based questions were prohibited.} per video for each category, ensuring diversity in entities and events.\footnote{Annotators were paid €100 for their work.} 
A manual review verified adherence to guidelines and semantic categories.
In the second phase (\emph{answer collection}), we used \textit{Prolific}\footnote{\url{https://www.prolific.com/}} to solve the task, targeting Italian-native participants with specific cultural criteria 
(aged $25$–$80$, born and raised in Italy, and Italian native speakers). Each annotator answered 12 out of 24 questions per video\footnote{Annotators were paid £7 per hour.}, focusing on detailed, visually grounded responses. Each question was answered by eight annotators 
to guarantee both accuracy and variability within the pool. \footnote{This choice is supported by \citet{vqa_evaluation_manos}, who show that up to 8 demonstrations provide a good balance between diversity, accuracy, and efficiency when talking about in-context learning with LLMs for VQA evaluation.}
Two semi-automatic validation checks were applied to the collected answers: ($1$) semantic consistency with the corresponding question, and ($2$) contradiction tests across answers in the same pool.\footnote{We found that 90.25\% of the 8-answer pools exhibit full agreement, as they do not contain any contradictions. The remaining 9.75\% was manually reviewed by an additional annotator to resolve inconsistencies. This shows agreement among annotators, that cannot be measured through standard annotator agreement metrics (e.g., Inter-Annotator Agreement) due to the open-ended nature of the task.} 
After validation, the dataset consists of $2{,}400$ questions, each paired with a pool of $8$ high-quality answers, for a total of $19{,}200$ validated responses. Through a post-processing of the lexicon, we made sure that the final 8-answer pools are lexical diverse.\footnote{The lexical overlap within the 8-answers pool is $21.90$\%, similarly to the overlap between pairs randomly extracted from each pool ($22.21$\%). In addition, the average \textit{Type-Token Ratio} (TTR) for content words within each pool is $0.55$. }

\textbf{Statement Collection.} 
As shown in Figure \ref{fig:MAIA_overview}, TSs are descriptive declarative sentences that accurately align with the visual content of videos.  TSs describe videos  from different semantic perspectives, according to MAIA semantic categories. 
TSs were generated using \textit{GPT-4o} (prompt in Figure \ref{fig:Prompt_StatementGen}A of the Appendix): for each question, it is given the 8 human generated answers and it is prompted to produce 8 TSs by combining the content of the question with the one of the corresponding answer. 
Again, post-processing techniques ensured high lexical variability within each pool reducing lexical overlap. FSs are incorrect descriptions automatically generated by prompting \textit{GPT-4o} (Figure \ref{fig:Prompt_StatementGen}B) and created by minimally modifying elements of a TS related to a reasoning category while maintaining the original sentence structure, thus forming minimal pairs. 
FSs were validated through two semi-automatic checks (\textit{GPT-4o}): 
(1) a structural verification to ensure that each FS was a minimal but incorrect variation consistent with its semantic category
, and (2) an NLI-based contradiction test to confirm that each FS contradicted its corresponding TS
. This process produced $19,200$ high-quality FSs aligned with their corresponding TSs.

\subsection{Reasoning Categories}
\label{sec:semcat}
We report  the reasoning categories  in MAIA.\footnote{More  details in Appendix \ref{subsec:Categories}.}  Figure \ref{fig:MAIA_overview} provides examples of aligned question, answer, TS, and FS.

\begin{description}[style=unboxed,leftmargin=0cm,noitemsep]
    \item[Causal.] {Focuses on questions about the cause or effect of an event. It provides a comprehensive test of a model's ability to infer and describe causality within events.  It can address either explicit (observable in the video) or implicit (inferred from the visible effect) causes/effects. }
    \item[Counterfactual.]{Focuses on hypothetical events that do not occur in the video but could happen under certain conditions. It tests a model's ability to reason about plausible scenarios grounded in the video's context.}
    \item[Implicit.] Involves questions about entities or events that are either not explicitly visible in the video (\textit{Total Implicit}) or no longer visible (\textit{Partial Implicit}), but can still be reasonably inferred. It evaluates a model's ability to deduce implicit details based on context.
    \item[Out-of-Scope.] Assumes the presence of entities or events not actually shown in the video, asking about properties of these nonexistent elements. It tests the model's ability to handle multimodal hallucinations and its tendency to make assertive, yet incorrect, responses.
    \item[Planning.] Inquires about the sequence of actions needed to achieve a specific goal related to the video. It assesses the model's ability to infer and plan the necessary steps based on the visual cues.
    \item[Sentiment.] Focuses on sentiment, mood, attitude, or emotions of characters towards other entities or events in the video. It evaluates the model's ability to recognize and identify the emotional cues.
    \item[Spatial.] Focuses on the location of entities in space, either applicable to the entire video (\textit{Total Spatial}) or specific moments and events (\textit{Partial Spatial}). It assesses the model’s ability to infer stable and time-dependent spatial relationships, determine relative positioning, and demonstrate grounding competencies.
    \item[Temporal.] Relies on when something happens, either in relation to other events (\textit{Partial Temporal}) or the duration of an event (\textit{Duration}). It evaluates the model's ability to infer temporal relationships, event sequences, and durations from visual content.
    \item[Uncertainty.] Arises when insufficient information is provided in the video to give a precise answer. It tests model's ability to handle situations with ambiguous or incomplete information, assessing its tendency to make assertive (rather than uncertain) responses.
\end{description}

\section{Experimental Setting}
\label{sec:exp_sett}

 We run several experiments to test modern VLMs on the MAIA benchmark in a zero-shot setting. To capture different VLM behaviors, strengths and limitations, we defined two tasks, aligned on the same datapoints: a multiple choice task, \textit{Visual Statement Verification} (VSV), and a generative task, \textit{Open-ended Visual Question Answering} (OEVQA).


\subsection{Task1: Visual Statement Verification}
\label{task1}
VSV is a multiple-choice task where a model is presented with a true-false statement pair related to a MAIA question (see section \ref{subsec: Dataset}) for a given video, and has to select the true option. The two statements are randomly assigned to two labels, A and B, and the model is asked to generate only the label. We chose the prompt through an extensive evaluation of $32$ variants ($16$ in Italian and $16$ in English), with the best-performing Italian prompt ultimately selected. 
Performance for VSV is measured with accuracy, i.e.,  the proportion of correctly selected true statements over the total statement pairs. 


\subsection{Task2: Open-ended VQA}
\label{task1}
OEVQA is a generative task, where models are tested on their ability to provide correct open-ended answers to a question related to video content. The model receives as input  a prompt question and a video, and is tasked with generating a correct answer. The prompt used in the experiments was selected as the best-performing among $10$ tested variants ($5$ in Italian and $5$ in English). 
Generated responses are then evaluated according to 
the following approaches
.

\begin{description}[style=unboxed,leftmargin=0cm,noitemsep]
\item [Similarity-based metrics.] It compares a  response against the pool of 8 reference answers available in MAIA.
We used five token-level metrics: \textit{ROUGE} \cite{lin-2004-rouge}, \textit{BLEU} \cite{10.3115/1073083.1073135}, \textit{BERT-Score} \cite{zhang2020bertscoreevaluatingtextgeneration}, \textit{METEOR} \cite{10.5555/1626355.1626389} and \textit{CIDEr} \cite{vedantam2015ciderconsensusbasedimagedescription}.

\item [LLM-as-a-judge.] While similarity metrics help rank VLMs, they fail in assessing answer correctness. To address this, we adopt an LLM-as-a-judge approach \cite{gu2025surveyllmasajudge}, using \textit{GPT-4o} to evaluate whether an answer is semantically consistent with at least one of the eight MAIA references, prioritizing meaning over surface-level structure (Appendix \ref{task_details}). Following \citet{bavaresco:llm24}, we validate this method on 100 samples: annotations by two human raters and \textit{GPT-4o} show strong agreement, with a Fleiss' Kappa of 0.82.

\end{description}

\begin{figure*}[h]
    \centering
    \begin{minipage}[b]{0.32\linewidth}
        \centering
        \includegraphics[width=\linewidth]{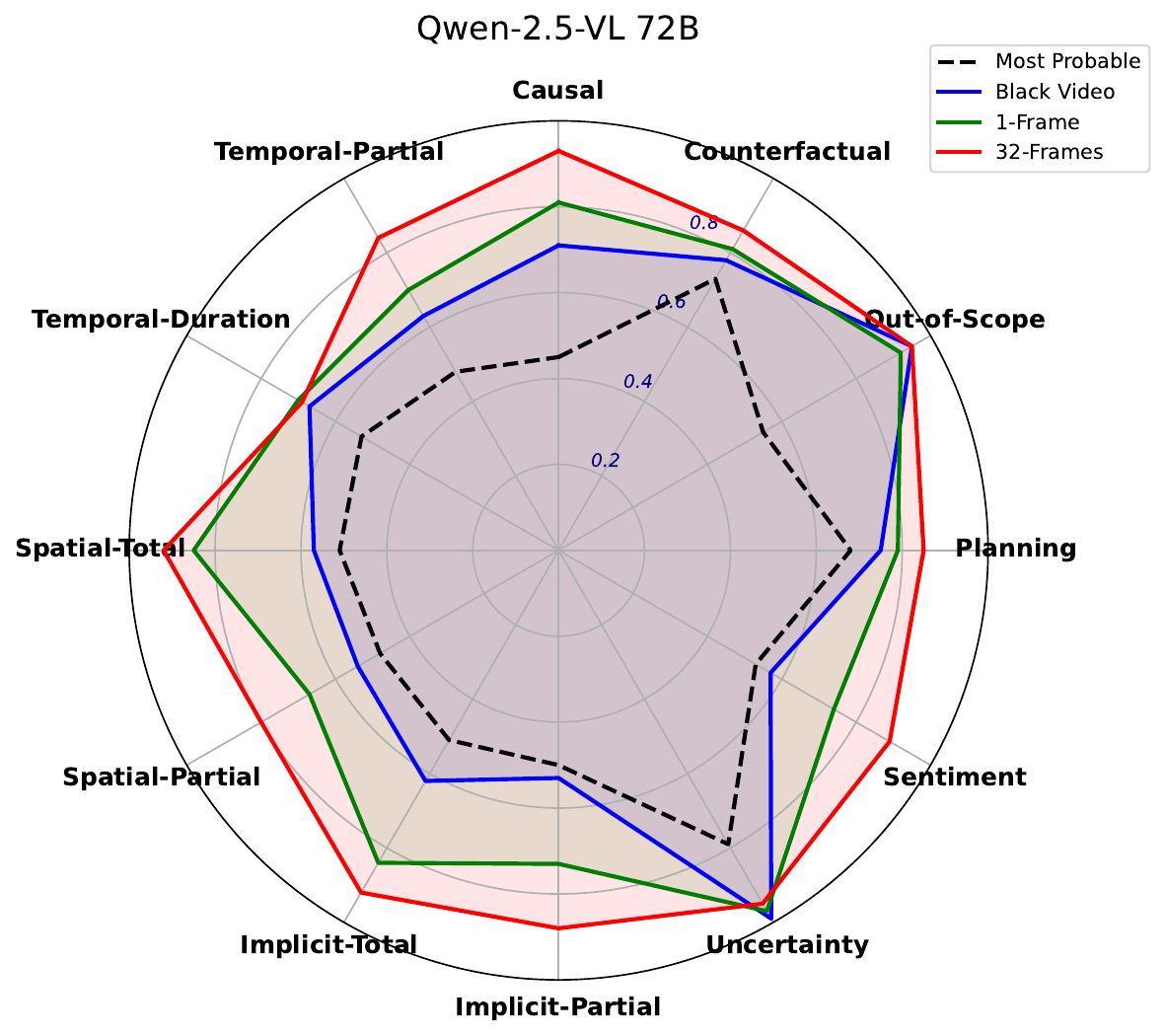}
        \caption*{(a) Task 1: VSV}
        \label{fig:spider_task1}
    \end{minipage}
    \hfill
    \begin{minipage}[b]{0.32\linewidth}
        \centering
        \includegraphics[width=\linewidth]{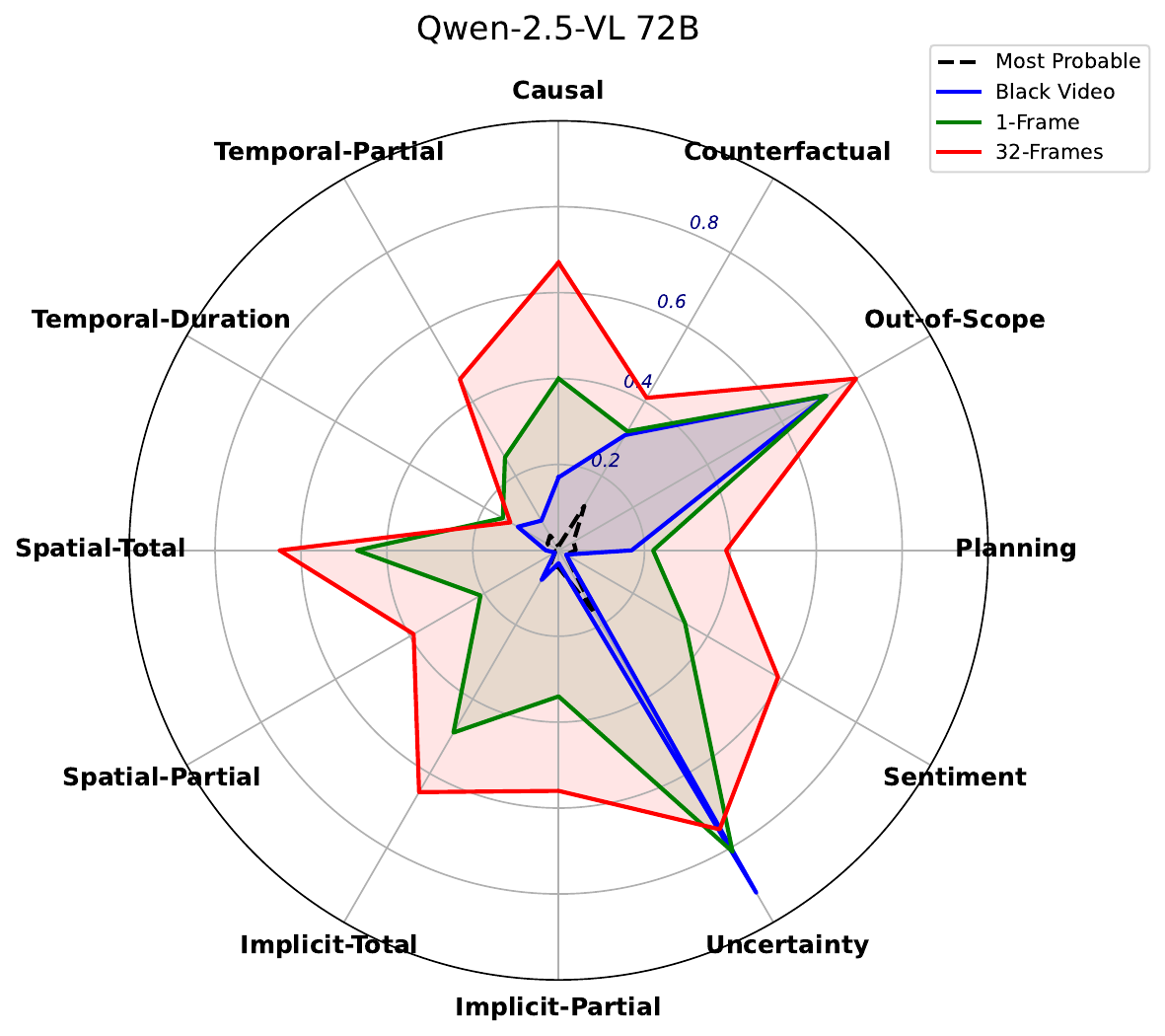}

        \caption*{(b) Task 1: VSV (pool-based)}
        \label{fig:spider_task1_consistency}
        
    \end{minipage}
    \hfill
    \begin{minipage}[b]{0.32\linewidth}
        \centering
        \includegraphics[width=\linewidth]{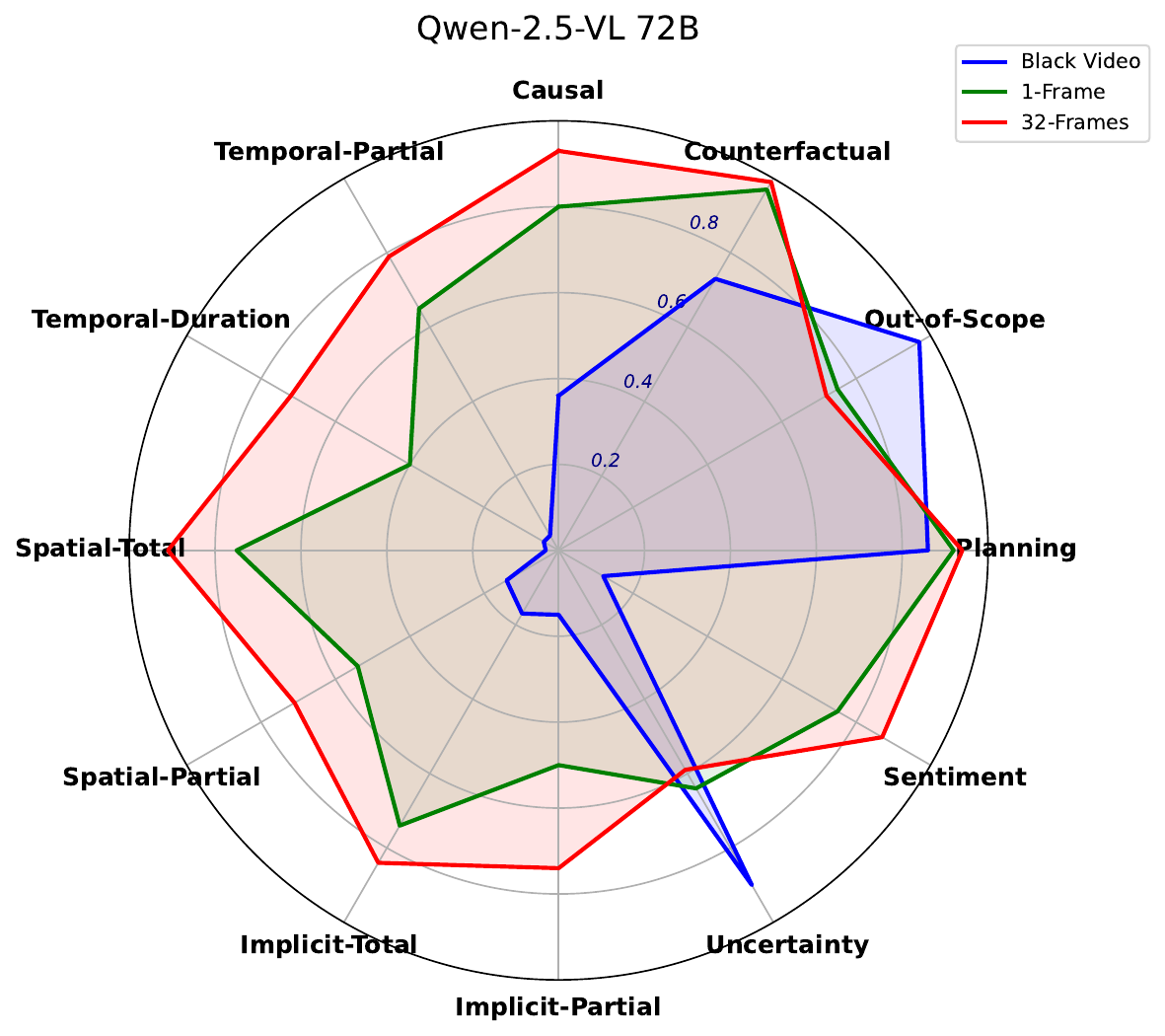}
        \caption*{(c) Task 2: OEVQA}
        \label{fig:spider_task2}
        
    \end{minipage}
    \caption{Fingerprint of \textit{Qwen2.5-VL} 72B through MAIA's reasoning categories: (a) illustrates model performance in NLU, Task 1, when TS-FS pairs are independent, while (b) reports performance on the same task when the model correctly identify all TS-FS pairs within each 8-item pool, thereby penalizing inconsistency; (c) visualizes the performance on NLG, Task 2.}
    \label{fig:spider}
\end{figure*}


\begin{table*}[]
\resizebox{\textwidth}{!}{%
\begin{tabular}{lc|c|c|c|c|c|c|c|cc|cc|cc}
\toprule
 & \textbf{Models} & \textbf{Avg.} & \textbf{Causal} & \textbf{Counterfactual}  & \textbf{Out-of-Scope} & \textbf{Planning} & \textbf{Sentiment}&  \textbf{Uncertainty} & \multicolumn{2}{c|}{\textbf{Implicit}} & \multicolumn{2}{c|}{\textbf{Spatial}} & \multicolumn{2}{c}{\textbf{Temporal}} \\
 &  &  &  &   &  &  &  & & \textit{Partial} & \textit{Total} & \textit{Partial} & \textit{Total} & \textit{Duration} & \textit{Partial} \\ 
 \midrule

\textbf{Unimodal} &  {} & 0.05 & 0.01 & 0.12   & 0.04 & 0.04 & 0.02 & 0.17 & 0.04 & 0.04 & 0.01 & 0.01 & 0.03 & 0.04 \\ 
\midrule

\multirow{6}{*}{\textbf{Black video}} & \textit{InternVL2} 8B  & 0.18 & 0.07 & 0.40   & 0.43 & 0.08 & 0.06 & 0.73 & 0.03 & 0.03 & 0.03 & 0.02 & 0.20 & 0.05 \\
  & \textit{InternVL3} 78B & \textbf{0.30}  & \textbf{0.35} & \textbf{0.44} & 0.04  & \textbf{0.28}  & \textbf{0.41} & \underline{\textbf{0.92}} & \textbf{0.10}   & \textbf{0.17} & 0.08 & \textbf{0.12} & \underline{\textbf{0.29}}  & \textbf{0.40} \\
 & \textit{Llava-Next-Video} 7B & 0.00 & 0.00 & 0.00   & 0.01 & 0.00 & 0.00 & 0.00 & 0.01 & 0.01 & 0.00 & 0.00 & 0.01 & 0.01 \\
 & \textit{Llava-oneVision} 7B & 0.14 & 0.01 & 0.04   & 0.59 & 0.04 & 0.00 & 0.83 & 0.04 & 0.04 & 0.01 & 0.01 & 0.06 & 0.01\\
 & \textit{Qwen-2.5-VL} 7B & 0.22 & 0.11 & 0.36   & 0.41 & 0.17 & 0.16 & 0.87 & 0.04 & 0.06 & \textbf{0.09} & 0.07 & 0.15 & 0.16 \\
  & \textit{Qwen-2.5-VL} 72B & 0.22 & 0.17 & 0.31 &  \textbf{0.72} & 0.17  & 0.02 & \underline{\textbf{0.92}} & 0.03  & 0.08  & 0.01 & 0.03 & 0.11  & 0.08 \\
\midrule

\multirow{5}{*}{\textbf{1-Frame}} & \textit{InternVL2} 8B & 0.28 & 0.32 & 0.41   & 0.32 & 0.11 & 0.33 & 0.47 & 0.17 & 0.33 & 0.15 & 0.39 & 0.11 & 0.21 \\
 & \textit{InternVL3} 78B & \textbf{0.42} & \textbf{0.53}  & \underline{\textbf{0.50}} & 0.46  &  \textbf{0.37} & \textbf{0.47} & \textbf{0.69} & 0.31 & \textbf{0.53} & \textbf{0.23} & 0.44  & 0.20   & \textbf{0.36}  \\
 & \textit{Llava-Next-Video} 7B  & 0.08 & 0.12 & 0.23   & 0.01 & 0.03 & 0.21 & 0.01 & 0.06 & 0.11 & 0.03 & 0.14 & 0.01 & 0.03 \\
 & \textit{Llava-oneVision} 7B & 0.32 & 0.35 & 0.29   & \textbf{0.64} & 0.14 & 0.32 & 0.65 & 0.21 & 0.35 & 0.12 & 0.36 & 0.15 & 0.20 \\
 & \textit{Qwen-2.5-VL} 7B & 0.36 & 0.36 & 0.28 & \textbf{0.64}   & 0.16 & 0.44 & \textbf{0.81}& 0.25 & 0.39 & 0.14 & 0.39 & \textbf{0.21} & 0.24  \\ 
  & \textit{Qwen-2.5-VL} 72B & 0.40 & 0.40  & 0.32  & 0.72  & 0.22 & 0.34 & \textbf{0.81} & \textbf{0.34} & 0.49  & 0.21 & \textbf{0.47} & 0.15 & 0.25 \\

 \midrule
 \midrule

\multirow{5}{*}{\textbf{32-Frames}} & \textit{InternVL2} 8B & {0.31} & 0.41 &  0.35   &  {0.38} &  {0.12} &  {0.42}&  {0.39} &  {0.28} &  {0.38} &  {0.18} &  {0.43} &  {0.11} &  {0.30} \\
 & \textit{InternVL3} 78B & 0.25  & 0.44 & \underline{\textbf{0.42}} & 0.14  & 0.21  & 0.22 & 0.20 & 0.24   & 0.36 & 0.17 & 0.20 & 0.06  & 0.26 \\
 &  \textit{Llava-Next-Video} 7B &  {0.03} &  0.04 &  {0.04}   &  {0.01} &  {0.03} &  {0.09} &  {0.00}&  {0.01} &  {0.06}&  {0.01} &  {0.04} &  {0.01} &  {0.01} \\
 &  \textit{Llava-oneVision} 7B &  {0.38} &  0.51 &  {0.21}   &  {0.61} &  {0.19} &  0.51 &  {0.45} &  {0.39} &  {0.47} &  {0.26} &  {0.48} &  {0.11} &  0.33 \\
 &  \textit{Qwen-2.5-VL} 7B &  0.44 &  0.53 &  {0.29}   &  0.63 &  0.23 &  0.50 &  0.67 &  0.43 &  0.56 &  0.28 &  0.55 &  \textbf{0.28} &  0.32\\ 
  & \textit{Qwen-2.5-VL} 72B & \underline{\textbf{0.54}} & \underline{\textbf{0.67}} & 0.41 &  \underline{\textbf{0.80}} & \underline{\textbf{0.39}}  & \underline{\textbf{0.59}} & \textbf{0.75} & \underline{\textbf{0.56}}  & \underline{\textbf{0.65}}  & \underline{\textbf{0.39}} & \underline{\textbf{0.65}} & 0.13  & \underline{\textbf{0.46}} \\
 \bottomrule
\end{tabular}%
}
\caption{\label{tab:task1_consistency}
    VSV (Task $1$): accuracy of correct pools (8/8) across reasoning categories, penalizing models' inconsistency.
  }
\end{table*}

\subsection{Baselines} 
We implemented three baselines for our tasks. 
\begin{description}
[style=unboxed,leftmargin=0cm,noitemsep]
    \item[Unimodal.] This baseline  applies only to Task 1 and  selects the most probable statement in a TS-FS pair.  It serves as a unimodal language baseline, reflecting the distributional biases of LLMs. Probabilities of TS and FS are first estimated on five open-weight LLMs that have shown good performance on a variety of Italian tasks \cite{magnini2025evalitallmbenchmarkinglargelanguage}: \textit{Llama-3.1} ($8$B-Instruct), \textit{LLaMAntino-2} ($7$B), \textit{LLaMAntino-3-ANITA} ($8$B-Instruct), \textit{Gemma} ($7$B) and \textit{Qwen2.5} ($7$B-Instruct). 
    For each TS-FS pair, we selected the item with the highest probability among the five models.
    \item[Black video.] It replaces MAIA videos with a fully black clip, used as a proxy for a no-video condition.
    This setup minimizes access to visual features, pushing the model to rely mainly on the language component, while the true unimodal evaluation remains the previous one. 
    \item[1-Frame.] This baseline considers only the first frame for each MAIA video, this way  reducing the capacity of a VLM to capture visual features and facing the one-frame “static appearance bias” \citep{lei-etal-2023-revealing}.  
\end{description}

\subsection{Vision-Language-Models}
    We benchmarked six recent VLMs sourced from the Hugging Face Hub, representing state-of-the-art approaches in Vision-Language tasks:\footnote{More details about VLMs are reported in Appendix \ref{sec:models_app} } \textit{InternVL2} (8B, \citet{internvl}), \textit{InternVL3} (78B, \citet{zhu2025internvl3exploringadvancedtraining}) \textit{LLaVA-NeXT-Video} (7B, \citet{llavanextvideo}), \textit{LLaVa-OneVision} (7B, \citet{llavaonevisioneasyvisualtask}), and \textit{Qwen2.5-VL} (both 7B and 72B, \citet{qwen2.5technicalreport}). All models accept a [video, text] pair as input, and uniformly sample 32 frames from the video.\footnote{During the experiments with \textit{InternVL3}, we found that, for about ten videos, $32$ frames exceeded the model’s context capacity, and in these cases we reduced them to $16$ to ensure proper processing.}


\label{ref:results}

\section{Results} \label{sec:results}
This section  reports the results obtained in our experiments for Task 1 and Task 2 independently.

\subsection{Results on Visual Statement Verification}
\label{ref:vvs}

Table~\ref{tab:task1_consistency} shows VLM accuracy across the three settings (black video, 1-Frame, 32-Frames) and reasoning categories when models consistently answers correctly (i.e., by choosing 8/8 the TSs within the 8-item pool). \textit{Qwen2.5-VL} 72B achieves the highest accuracy in the 32-Frames setup with an average score of $0.54$, marking a 14-point improvement over the correspondent 1-Frame setting. \textit{Llava-Next-Video} shows the weakest performance across all three configurations, likely due to its underlying Vicuna-7B LLM, weakly trained in Italian.
All other models outperform the unimodal baseline ($0.05$):  in the 32-Frames setting, gains go from +$26$ points (\textit{InternVL2}) to +$49$ (\textit{Qwen2.5-VL} 72B), confirming the use of visual cues to counteract language-driven biases. Notably, \textit{InternVL3} 78B, despite being a high-performing model, exhibits an opposite trend in the 32-Frames setting, where its accuracy drops by $0.05$ and $0.17$ compared to the Black Video and 1-Frame configurations, instead of improving as observed for the other models.\footnote{One possible explanation is that the model has not been trained to process 32-frame inputs at the resolution we provide.}\\
As a complement, Figure~\ref{fig:spider} helps visualizing the comparison across reasoning categories.  Here we report results only for \textit{Qwen-2.5-VL} 72B, our best model, while similar figures for the other models are in  Appendix~\ref{sec:appendixB} (Figure \ref{fig:spider_task1}, \ref{fig:spider_task1_cons} and \ref{fig:spider_task2}). The most difficult categories are \textsc{Counterfactual}, \textsc{Planning}, \textsc{Implicit} \textit{Partial},  \textsc{Spatial} \textit{Partial}, and \textsc{Temporal} \textit{Duration}. In addition, \textsc{Causal}, \textsc{Sentiment}, \textsc{Implicit} \textit{Partial} and \textit{Total}, and \textsc{Spatial} \textit{Partial} and \textit{Total} are the categories for which the model profits the most from visual clues, by leveraging the broader visual window provided in the 32-Frames setting, as shown by the larger area covered by the red curve compared to the green and blue ones.
\paragraph{Models' consistency.} Figure~\ref{fig:spider}a vs.\ ~\ref{fig:spider}b highlights the role of the consistency check in MAIA by using a pool of 8 TS-FS pairs.  When pairs are considered independently, as it is usually done in VLM benchmarks, the model's performance increases 
significantly, showing that it relies on spurious correlation, effects that is strongly mitigated by the MAIA's severe evaluation regime. In our case, such trend is even much more visible in the black video setting than in the 32-Frames one with a gain of +$47$ vs.\ +$34$ for \textit{Qwen2.5-VL} 72B (see Appendix, Table \ref{tab:task1_res_app}).
By systematically comparing models' performance in the independent and pool-based settings, we see that this is a general finding across models. Moreover, we find that \textit{Qwen-2.5-VL} 72B is not only the best-performing model, but also the most consistent, with a lower drop in the pool-based accuracy (Table \ref{tab:task1_consistency}) with reference to the indipendent one (Table \ref{tab:task1_res_app}), particularly in the 32-Frames setting (i.e., 34 points).

\subsection{Results on Open-ended Generation}
\label{ref:Open_gen}

\begin{table*}[]
\resizebox{\textwidth}{!}{%
\begin{tabular}{lc|c|c|c|c|c|c|c|cc|cc|cc}
\toprule
 & \textbf{Models} & \textbf{Avg.} & \textbf{Causal} & \textbf{Counterfactual}  & \textbf{Out-of-Scope} & \textbf{Planning} & \textbf{Sentiment}&  \textbf{Uncertainty} & \multicolumn{2}{c|}{\textbf{Implicit}} & \multicolumn{2}{c|}{\textbf{Spatial}} & \multicolumn{2}{c}{\textbf{Temporal}} \\
 &  &  &  &   &  &  &  & & \textit{Partial} & \textit{Total} & \textit{Partial} & \textit{Total} & \textit{Duration} & \textit{Partial} \\ 
 \midrule

\multirow{5}{*}{\textbf{Black video}} & \textit{InternVL2} 8B & 0.37 & 0.42 & 0.60  & 0.30 & 0.68 & 0.43& 0.08 & 0.21 & 0.23  & 0.36 & 0.25 & \textbf{0.55} & 0.25 \\
& \textit{InternVL3} 78B & \textbf{0.52} & \textbf{0.77}  & \textbf{0.96}  &  0.00 & 0.83  & \textbf{0.74} & 0.09 & \textbf{0.33} & \textbf{0.35}  & \textbf{0.50} & \textbf{0.54} & 0.49 & \textbf{0.62}    \\
 & \textit{Llava-Next-Video} 7B & 0.27 & 0.43 & 0.47   & 0.30 & 0.40 & 0.33& 0.51 & 0.21 & 0.12 & 0.16 & 0.04 & 0.12 & 0.16 \\
 & \textit{Llava-oneVision} 7B & 0.40 & 0.66 & 0.68   & 0.29 & 0.60 & 0.60& 0.28& 0.26 & 0.24 & 0.36 & 0.18 & 0.37 & 0.23 \\
 & \textit{Qwen-2.5-VL} 7B & 0.35 & 0.36 & 0.69  & 0.08 & 0.54 & \textbf{0.74}& 0.23& 0.24 & 0.19 & 0.30 & 0.19 & 0.41 & 0.20 \\
   & \textit{Qwen-2.5-VL} 72B & 0.38  &  0.36 &  0.73  & \underline{\textbf{0.97}}  & \textbf{0.86} & 0.12 & \underline{\textbf{0.90}} & 0.15 & 0.17 & 0.14 & 0.03 & 0.04  & 0.04 \\
 \midrule
 
\multirow{5}{*}{\textbf{1-Frame}} & \textit{InternVL2} 8B & 0.44 & 0.57 & 0.65   & 0.21 & 0.65 & 0.60& 0.10& 0.33 & 0.38 & 0.39 & 0.47 & 0.53 & 0.35  \\
& \textit{InternVL3} 78B & 0.68 & \textbf{0.88} & \textbf{0.97} & 0.44  & 0.91  & \textbf{0.83} & 0.56 & \textbf{0.55}  & 0.70 & \textbf{0.57} & \textbf{0.77} & 0.44  & 0.56  \\
 & \textit{Llava-Next-Video} 7B & 0.32 & 0.30 & 0.56   & 0.20 & 0.46 & 0.60& 0.29& 0.18 & 0.32& 0.24 & 0.27 & 0.20 & 0.19 \\
 & \textit{Llava-oneVision} 7B & 0.50 & 0.59 & 0.78  & 0.15 & 0.66 & 0.74  & 0.37& 0.39 & 0.54& 0.39 & 0.51 & 0.54 & 0.33 \\
 & \textit{Qwen-2.5-VL} 7B & 0.51 & 0.56 & 0.76   & 0.32 & 0.60 & 0.79 & 0.40 & 0.35 & 0.50 & 0.38 & 0.53 & \textbf{0.55} & 0.38 \\
   & \textit{Qwen-2.5-VL} 72B & \textbf{0.70}  & 0.80 &  \textbf{0.97}  & \underline{\textbf{0.75}}  & \textbf{0.92} & 0.75 & \textbf{0.64} & 0.50 & \textbf{0.74} & 0.54 & 0.75 & 0.40  & \textbf{0.65} \\
 \midrule
  \midrule
 
\multirow{5}{*}{\textbf{32-Frames}} & \textit{InternVL2} 8B & 0.49 & 0.54 &  0.68   &  0.28 &  0.64 &  0.62&  0.11 &  0.45  &  0.48&  0.47 &  0.51 &  0.57 &  0.46 \\
& \textit{InternVL3} 78B & 0.77  & 0.86 & 0.96  & 0.54  & 0.77  & 0.85 & \textbf{0.64} & 0.73  &  0.81 & 0.50 & 0.54  & 0.49   & 0.62 \\
 &  \textit{Llava-Next-Video} 7B &  0.33 &  0.37 &  0.38   &  0.16 &  0.42 &  0.48&  0.27 &  0.24 &  0.37 &  0.29 &  0.39 &  0.32 &  0.29 \\
 &  \textit{Llava-oneVision} 7B &  0.53 &  0.67 &  0.79  &  0.11 &  0.65 &  0.79&  0.23 &  0.55 &  0.56 &  0.51 &  0.62 &  0.40 &  0.46 \\
 &  \textit{Qwen-2.5-VL} 7B &  0.61 &  0.71 &  0.80   &  0.43 &  0.60 &  0.85 &  0.55&  0.55 &  0.60 &  0.50 &  0.70 &  0.54 &  0.53 \\ 
   & \textit{Qwen-2.5-VL} 72B & \underline{\textbf{0.81}}  & \underline{\textbf{0.93}}  &  \underline{\textbf{0.99}}  & \textbf{0.72} & \underline{\textbf{0.94}} &  \underline{\textbf{0.87}} & 0.59  & \underline{\textbf{0.74}} &  \underline{\textbf{0.84}} & \underline{\textbf{0.71}} & \underline{\textbf{0.91}}  & \underline{\textbf{0.72}} & 
   \underline{\textbf{0.79}} \\
  
 \bottomrule
 
\end{tabular}%
}
 \caption{\label{tab:Task2_llm_as_judge}
    OEVQA (Task $2$): accuracy of correct answers with LLM-as-a-judge.
  }
\end{table*}

\begin{table*}[h]
\resizebox{\textwidth}{!}{%
\begin{tabular}{lc|c|c|c|c|c|c|c|cc|cc|cc}
\toprule
 & \textbf{Model} & \textbf{Avg.} & \textbf{Causal} & \textbf{Counterfactual}  & \textbf{Out-of-Scope} & \textbf{Planning} & \textbf{Sentiment}&  \textbf{Uncertainty} & \multicolumn{2}{c|}{\textbf{Implicit}} & \multicolumn{2}{c|}{\textbf{Spatial}} & \multicolumn{2}{c}{\textbf{Temporal}} \\
 &  &  &  &   &  &  &  & & \textit{Partial} & \textit{Total} & \textit{Partial} & \textit{Total} & \textit{Duration} & \textit{Partial} \\ 
 \midrule

\multirow{1}{*}{\textbf{Black video}} 
 & \textit{Qwen-2.5-VL} 72B & 0.18 & 0.08 & 0.30 & \underline{\textbf{0.69}} & 0.15 &  0.01 & \underline{\textbf{0.84}} & 0.02 & 0.02 & 0.00 & 0.00 & 0.00 & 0.00 \\
 \midrule
 
\multirow{1}{*}{\textbf{1-Frame}} 
 & \textit{Qwen-2.5-VL} 72B & 0.33 & 0.39 & 0.32   & 0.59 & 0.20 & 0.32 & 0.53 & 0.27 & 0.43  & 0.15 & 0.42 & 0.09 & 0.24 \\ 
 \midrule
  \midrule
 
\multirow{1}{*}{\textbf{32-Frames}}
 &  \textit{Qwen-2.5-VL} 72B &  \underline{\textbf{0.47}} &  \underline{\textbf{0.64}} &  \underline{\textbf{0.41}}   &  0.61 &  \underline{\textbf{0.37}} &  \underline{\textbf{0.56}} &  0.49 &  \underline{\textbf{0.48}} &  \underline{\textbf{0.58}} &  \underline{\textbf{0.35}} &  \underline{\textbf{0.62}} &  \underline{\textbf{0.10}} &  \underline{\textbf{0.40}} \\
 
 \bottomrule
\end{tabular}%
}
 \caption{\label{tab:tasks_consistency_threshold8}
    Aggregate accuracy on Task $1$ (NLU) and Task $2$ (NLG) (consistency and robustness) on \textit{Qwen-2.5-VL} 72B across reasoning categories.
  }
\end{table*}

Table~\ref{tab:Task2_llm_as_judge} reports the results on Task 2. The best performing model  is again \textit{Qwen2.5-VL} 72B, reaching $0.81$ accuracy in the 32-Frames setting. Unlike in Task $1$, here \textit{InternVL3} 78B shows a positive incremental trend, with performance progressively improving from Black Video to 1-Frame and reaching its best results with 32 frames. In the latter configuration, across models, the hardest categories are \textsc{Uncertainty}, \textsc{Out-of-scope}, \textsc{Implicit} \textit{Partial}, \textsc{Temporal} \textit{Duration} and \textit{Partial}, and \textsc{Spatial} \textit{Partial}, while \textsc{Counterfactual} and \textsc{Planning} appear less challenging in this context. This overall tendency is also confirmed for our best model.
In particular, Figure~\ref{fig:spider}c shows that the model benefits from 32 frames in all categories but \textsc{Out-of-scope} and \textsc{Uncertainty}, where it instead excels with black videos, though in most other cases performance in this setting is poor. Surprisingly, for \textsc{Counterfactual} and \textsc{Planning}, a relatively high accuracy is obtained already with black videos. Interestingly, differences also emerge when comparing the 1-Frame vs.\ 32-Frames settings. In the majority of cases, the 1-Frame is not enough, while 32 frames increase performance. This difference is less pronounced or does not hold for \textsc{Planning}, \textsc{Counterfactual} and \textsc{Uncertainty}.
For example, in the \textsc{Implicit} \textit{Total} category, the 32-Frames model answers correctly to \textit{How does the vehicle move?}\footnote{Referring to a pirate ship in the amusement park} with \textit{The vehicle moves in a swinging way, with back-and-forth movements}, while the 1-Frame setting hallucinates with \textit{The vehicle moves slowly along the amusement park route}.

\begin{table}[t!]
  \centering
  \resizebox{\linewidth}{!}{%
  \renewcommand{\arraystretch}{1.2}
  \setlength{\tabcolsep}{4pt}
  \begin{tabular}{l|c|c|c|c|cc} 
    \toprule
       \textbf{Models} & ROUGE & BertScore & BLEU & METEOR & CIDEr \\ 
    \midrule
    \midrule
    \textit{InternVL2} & 0.61 & \textbf{0.84} & 0.38 & 0.59 & \textbf{1.18}\\
    \textit{InternVL3} 78B & 0.50 & 0.80  & 0.26 & 0.47   &  0.67  \\
    \textit{LLaVa-NeXT-Video} & 0.46 & 0.79 & 0.21  & 0.45 & 0.65\\ 
    \textit{LLava-oneVision} & 0.58 & 0.83  & \textbf{0.40} & 0.55 & 1.08\\ 
    \textit{Qwen-2.5-VL} & 0.58 & 0.83 & 0.38 & 0.61 & 0.98\\ 
    \textit{Qwen-2.5-VL} 72B & \textbf{0.62} & \textbf{0.84} & 0.37 & \textbf{0.65} & 1.07 \\
    \bottomrule
  \end{tabular}%
  }
  \caption{{\label{tab:similarity_metrics_vqa}}VLM performance (32-Frames setting) for OEVQA (Task 2) according to similarity-based metrics.}

\end{table}

\noindent Table \ref{tab:similarity_metrics_vqa} highlights how similarity-based metrics (e.g., ROUGE, BLEU) often do not align with semantic correctness. \textit{InternVL2} scores highest on surface-level similarity (e.g., BERTScore: $0.84$, CIDEr: $1.18$) -- as well as \textit{Qwen2.5-VL} 72B --  but lower when considering the LLM-as-a-judge metric ($0.49$), while \textit{Qwen2.5-VL} 72B and \textit{InternVL3} 78B offer more (semantic) accurate answers. \textit{LLaVA-NeXT-Video} underperforms across all evaluations.

\subsection{Discussion}
\label{subsec:discussion}

The results presented for Task 1 and 2 clearly highlight two key findings. 
First, within each task, the role of the information extracted from videos is unequally distributed across reasoning categories. The star-shaped Figure~\ref{fig:spider}b illustrates such differences: the star's picks highlight the categories that profit from the visual input the most: \textsc{Spatial} \textit{Total}, \textsc{Implicit} \textit{Total}, \textsc{Causal}, \textsc{Out-of-Scope}, and \textsc{Uncertainty}.  On the other hand, and quite surprisingly, \textit{Qwen2.5-VL} 72B handles \textsc{Out-of-Scope} and \textsc{Uncertainty} better when provided the black videos, which are expected to be uninformative, than with the full 32 frames. This calls for a deeper analysis of the reason behind such a result, as the more visual context the model receives, the more it hallucinates, reducing \textsc{Out-of-Scope} scores, and becomes overly assertive, reducing \textsc{Uncertainty} scores.
Overall, this shows that MAIA categories are extremely useful to factorize the contribution of the visual vs. linguistic components of VLMs, favoring a more nuanced analysis of their actual abilities. Finally, among the most challenging categories -- discussed in \ref{ref:vvs} and \ref{ref:Open_gen} -- several (e.g., \textsc{Spatial} \textit{Partial}, \textsc{Implicit} \textit{Partial}, and \textsc{Temporal} \textit{Duration}) share a temporal dimension, reinforcing the evidence that temporal reasoning still remains a fundamental issue of current models.  \\A second noteworthy fact is the effect of the task design. 
By comparing Table~\ref{tab:task1_consistency} and Table~\ref{tab:Task2_llm_as_judge}, we see that overall accuracy is higher in Task $2$ than in Task $1$, even though the underlying information the model has to exploit to perform the tasks is the same, given the alignment between their data points. Such an increase is found across models: +$0.26$ increase on average -- in line with the Generative AI Paradox -- they are better at generating than at understanding 
text~\cite{west2024the}.  Figure~\ref{fig:spider}b vs.\ Figure~\ref{fig:spider}c illustrates such a difference for \textit{Qwen2.5-VL} 72B.
Interestingly, the difference is more pronounced for some categories, as shown by the fact that Figure~\ref{fig:spider}c no longer has the shape of a star; for instance, \textsc{Planning}, \textsc{Counterfactual}, \textsc{Spatial} \textit{Partial} and \textsc{Temporal} \textit{Duration} improve the most. On the other hand, \textsc{Uncertainty} and \textsc{Out-of-scope} accuracy drops in the NLG task; from a qualitative analysis, we saw that this is mostly due to the generation of hallucinations (Appendix \ref{sec:app_mhallucinations}).

However, models' size proved to be a crucial factor: larger models consistently achieve higher accuracy, even within the same family, suggesting that scaling provides apparent advantages on individual tasks, although this picture changes when considering MAIA’s broader scope (as discussed in the next section).

\section{Aggregating Understanding and Generation}
\label{sec:Understanding-Generation}
A distinctive feature of MAIA is the alignment between the VSV and OEVQA tasks: both are grounded in the same question and the same video. While Section \ref{sec:results} reported their results separately, we now combine them into a unified evaluation framework that jointly tests comprehension and generation abilities in VLMs.
We introduce \textit{Aggregate Accuracy} (\textit{Agg-Acc}), a metric rewarding models that: (i) consistently select the correct statement (TS) over all 8 TS-FS pairs in Task 1, and (ii) generate a correct answer to the same question in Task 2, according to our LLM-as-a-judge evaluation. The idea is to capture overlapping abilities (i.e.\ knowledge and reasoning) for the two aligned tasks, evaluating models' robustness. \textit{Aggregate Accuracy} is defined as follow:
\[
Agg\text{-}Acc(M, q) = 
\begin{cases} 
1 & \text{if } \forall (TS, FS) \in S_q, \\
  & a_M(TS, FS) = TS \\
  & \text{and } a_M(q) \text{ is correct} \\
0 & \text{otherwise}
\end{cases}
\]

\noindent
where $M$ is the model, $q$ a question, $S_q$ the set of TS-FS pairs, and $a_M$ the model's answers. Intuitively, given a question $q$ on a video $v$, we reward the ability of a VLM to both select a correct answer $TF$ from a $TF-FS$ pair related to $q$, and to generate a correct answer $a$ to $q$, as discussed in Section~\ref{sec:intro} when commenting the question \textit{What is the cake made of?} and the aligned data of the NLU and NLG tasks in Figure~\ref{fig:maia_structure}.

Table \ref{tab:tasks_consistency_threshold8} reports \textit{Agg-Acc} for our best model \textit{Qwen2.5-VL} 72B: $0.47$ with 32-frames, $0.33$ with 1-frame, and only $0.18$ with black video; this shows the more challenging nature of the aggregate task, and that a single frame is not that robust.
\textsc{Planning}, \textsc{Spatial} \textit{Partial}, and \textsc{Temporal} \textit{Duration} remain challenging even with 32-Frames, despite their notable improvements from Task~$1$ to Task~$2$.
 Results confirm that MAIA’s aggregated understanding\&generation task creates a harder benchmark, 
 laying the ground toward a more objective VLM evaluation, even when considering larger models that apparently achieve high accuracy and appear to perform well.

\section{Conclusion}
We introduce MAIA, a benchmark designed for fine-grained investigation of the reasoning abilities of VLMs on videos. MAIA has two aligned tasks: a visual statement verification task (NLU), and a open-ended visual question answering task (NLG), both on the same set of video related questions.  First, we provided a in-depth analysis of the two tasks independently, showing the importance of evaluating model with answer-pools to account for model consistency. 
Then, we couple comprehension and generation in an aggregated evaluation framework, arguing that the aggregated "all-in-one" understanding\&generation task is a challenging and natural 
setting toward a more objective VLM evaluation, as it reveals inconsistencies within the same task and a lack of robustness across aligned tasks, even in larger models. As for the future, it would be interesting  to see whether our framework promote models that undergo learning paradigms tightly integrating these two capabilities, as in~\citet{gul-artzi-2024-cogen}.

\section*{Limitations and Future Directions}
We acknowledge that the number of videos in MAIA may seem relatively small, with only 100 samples. However, their combination with 12 reasoning categories results in 19,200 samples for Task 1 and an equal number of question-answer pairs, providing a robust evaluation set, not intended for any kind of training. Still, these 100 videos constitute only the initial core of a broader evaluation framework planned for future development.
Moreover, we are aware that our most probable baseline, constructed using probabilities derived from our set of LLM's logits, poses a limitation that we intend to address in future work. We plan to compare each VLM with its corresponding LLM to obtain more reliable results for proper comparisons and analyses of potential statistical biases. Regarding Task $2$, we aim to dive deep into the comparison between the similarity metrics used and the emerging topic of LLMs as judges. Additionally, we intend to further investigate this latter approach to assess its actual validity as a reliable evaluation method. This direction will help us refine accuracy metrics, ultimately enhancing our ability to rigorously test model robustness and consistency across our two specular tasks. 
Furthermore, we acknowledge that
our evaluation did not include large-scale proprietary models (e.g., ChatGPT). Our focus was primarily on testing the performance of open-source
and easily exploitable language models to provide
a comprehensive overview of their capabilities on
the benchmark. 
Finally, we are also aware that our benchmark has not yet been compared with existing ones to assess its relative difficulty and challenge level. This limitation comes from the lack of comparable resources in Italian. As a future direction, we plan to translate MAIA into English and replicate our experiments, enabling more meaningful comparisons with similar English-language benchmarks.

\section*{Acknowledgments}
This work has been carried out while Davide Testa was enrolled in the Italian National Doctorate on Artificial Intelligence run by Sapienza University of Rome in collaboration with Fondazione Bruno Kessler (FBK). Giovanni Bonetta and Bernardo Magnini were supported by the PNRR MUR project \href{https://fondazione-fair.it/}{PE0000013-FAIR} (Spoke 2). Alessandro Lenci was supported by the PNRR MUR project \href{https://fondazione-fair.it/}{PE0000013-FAIR} (Spoke 1). Alessio Miaschi was supported by the PNRR MUR project \href{https://fondazione-fair.it/}{PE0000013-FAIR} (Spoke 5). Lucia Passaro was supported by the EU EIC project EMERGE (Grant No. 101070918).
Alessandro Bondielli was supported by the PNRR MUR project \href{https://fondazione-fair.it/}{PE0000013-FAIR} (Spoke 1), funded by the European Commission under the NextGeneration EU programme and by the the Italian Ministry of University and Research (MUR) in the framework of the PON 2014-2021 ``Research and Innovation" resources – Innovation Action - DM MUR 1062/2021 - Title of the Research: ``Modelli semantici multimodali per l’industria 4.0 e le digital humanities.''




\appendix

\section{MAIA:Benchmark Details}
\label{sec:appendixA}

\subsection{Semantic Categories Definition}
\label{subsec:Categories}

We report here the definition of the twelve reasoning categories included in MAIA.
\begin{description}
[style=unboxed,leftmargin=0cm,noitemsep]

    \item[CAUSAL.] This category\footnote{Note that in MAIA there are four macro-categories with two fine-grained specifications (i.e., subcategories). The only exception is the \textit{Causal} category, in which explicit and implicit items are equally represented (100 each). However, we do not consider them subcategories in the same way as the others, since in those cases, the subcategories express entirely different aspects of the same domain.} includes two subtypes: \textit{Implicit Causal} and \textit{Explicit Causal}, both aimed at reasoning about the causes or effects of events depicted in the video. Thus, it includes reasoning tasks involving both visible and inferred causal relationships, offering a comprehensive test of a model's ability to infer and describe causality within events.
    
    \begin{itemize}
        \item \underline{Implicit Causal}: This type of question targets the inferred cause of an event, object, or human action visible in the video. The focus is on an implicit cause that cannot be directly observed but must be deduced from the effect presented in the scene. Typical responses involve a logical inference explaining the implicit cause behind the visible effect.\\
        Example: Suppose a video shows a person at home grabbing an umbrella while going out.\\
        
        \begin{quote}
            Italian:\\
    Q: \textit{Per quale motivo la persona prende l'ombrello?}\\
    A: \textit{Perchè potrebbe piovere fuori.}\\
    
        \end{quote}
        
        \begin{quote}
            English: \\
    Q: \textit{Why does the person take the umbrella?}\\
    A: \textit{Because it might be raining outside.}\\
    
        \end{quote}
    In this example, the action of grabbing the umbrella is visible, but the reason (bad weather) is not explicit in the video and must be inferred.
    
        \item \underline{Explicit Causal}: This type of question addresses direct cause-and-effect relationships visible within the video. The focus is on identifying a specific event, object, or human action (the cause) that led to another event, situation, or state (the effect) or vice versa. Typical responses clearly describe either the cause or the effect based on what is directly observable in the video.\\
        Example: Suppose a video shows an angry person throwing a glass on the floor, which subsequently shatters.\\
        
        \begin{quote}
           Italian:\\
    Q: \textit{Perchè il bicchiere si è rotto?}\\
    A: \textit{Perchè la persona lo ha gettato a terra}.\\
        \end{quote}
        \begin{quote}
            English:\\
    Q: \textit{Why did the glass break?}\\
    A: \textit{Because the person threw it on the ground}.\\
    
        \end{quote}
        Here, both the cause (throwing the glass) and the effect (the glass breaking) are visible in the video and can be used to provide a direct response.
    \end{itemize}
    
    \item[COUNTERFACTUAL.] This category focuses on questions about hypothetical scenarios that do not actually occur in the video but could take place under specific conditions. These questions explore the consequences of an event or situation that might happen in the video if a certain condition were met. A key requirement is that the hypothetical condition must be based on entities or events visible in the video.
    Consequently, this category tests a model's ability to reason about hypothetical scenarios grounded in the context of the video while deriving logical and plausible outcomes from such scenarios.\\
    Example: Suppose a video shows an outdoor concert.
    \begin{quote}
            Italian:\\
    Q: \textit{Cosa succederebbe al concerto se arrivasse un forte temporale?}\\
    A: \textit{Il concerto verrebbe interrotto all'istante.}\\
    
        \end{quote}
        
    \begin{quote}
            English:\\
    Q: \textit{What would happen to the concert if a violent thunderstorm started?}\\
    A: \textit{The concert would be immediately interrupted.}\\
    
        \end{quote}
        
    In this example, the focus of the question (the concert) is visible in the video, while the condition (a violent thunderstorm) is not. The consequence (the concert being interrupted) is not shown in the video but can be reasonably inferred
    .
    \item[IMPLICIT.] The implicit category includes questions about entities, events, or their attributes that are not explicitly visible in the video. However, their presence or properties can be reasonably inferred from the context. This category evaluates a model’s ability to infer implicit details based on context, whether the target information was never shown or was previously visible but later obscured.

    \begin{itemize}
        \item \underline{Total Implicit}:
    These questions focus on entities or events that are never directly visible in the video but can be inferred from observable details. A typical answer provides the requested information based on logical inference.\\
    Example: Suppose a video shows the interior of a house, and suddenly the front door opens, revealing a person soaking wet with a dripping closed umbrella.
    
    \begin{quote}
       Italian: \\     
    Q: \textit{Che tempo fa fuori?}\\
    A: \textit{Piove molto forte.}\\
    
        \end{quote}

    \begin{quote}
       English: \\     
    Q: \textit{What’s the weather like outside?}\\
    A: \textit{It’s raining heavily.}\\
    
        \end{quote}
    In this case, the focus of the question (the weather outside) is not visible at any point in the video. However, details such as the wet person and dripping umbrella allow for a reasonably confident inference (heavy rain).
    
        \item \underline{Partial Implicit}:
    These questions address entities or events that were visible earlier in the video but are no longer visible due to a shift in the scene or because they have moved out of the frame.\\
    Example: Suppose a video shows a man placing a pen in a drawer and then closing it.
    
    \begin{quote}
        Italian: \\   
    Q: \textit{Dove si trova la penna?}\\
    A: \textit{La penna è nel cassetto.}\\
    
        \end{quote}
    
    \begin{quote}
        English: \\   
    Q: \textit{Where is the pen?}\\
    A: \textit{The pen is inside the drawer.}\\
    
        \end{quote}
    In this example, the focus of the question (the pen) is no longer visible in the video. However, earlier information (the man placing the pen in the drawer) allows for a logical and confident answer (the pen is in the drawer).
    \end{itemize}


    \item[OUT-OF-SCOPE.] Such a category involves questions about entities or events that are not present in the video at all, asking for properties or details about these non-existent entities or events. A typical response to an out-of-scope question is a negation, stating that the entity or event in question is not present. This category tests the ability of a model to identify and handle irrelevant or non-existent entities within the video content, appropriately responding with a negation when the requested object or event is absent. Thus, it represents an indirect way to test the models on possible multimodal hallucinations and their tendency to be assertive in their responses.\\
    Example: Suppose a video shows a dog and its owner playing in the park, but there are no cars in the scene.
    \begin{quote}
            Italian:\\
    Q: \textit{Di che colore è la macchina?}\\
    A: \textit{Non ci sono auto (nella scena).}\\
    
        \end{quote}
    \begin{quote}
            English:\\
    Q: \textit{What color is the car?}\\
    A: \textit{There is no car (in the scene).}\\
    
        \end{quote}
    In this example, the focus of the question (the car) is not physically present in the video, nor can its presence be reasonably inferred. When trying to answer the question, no useful information about a car can be found, and the expected response would be a negation, such as "There is no car." 


    \item[PLANNING.] This category involves questions that request the actions needed to achieve a specific goal related to the video. The typical response to a planning question is a sequence of actions that someone should perform, based on the situation presented in the video, in order to reach the desired outcome. Such a category assesses the model's ability to infer and plan the necessary steps to accomplish a goal based on the visual cues provided in the video.\\
    Example: Suppose a video shows a dog and its owner playing with a ball in a park, and the owner throws the ball onto a bench.
    
    \begin{quote}
            Italian:\\
    Q: \textit{Cosa dovrebbe fare il cane per continuare a giocare col padrone?}\\
    A: \textit{Dovrebbe correre verso la palla, saltare sulla pacchina, prendere la palla e riportarla al padrone.}\\
        \end{quote}

    \begin{quote}
            English:\\
    Q: \textit{What should the dog do to continue playing with its owner?}\\
    A: \textit{The dog should run toward the ball, jump onto the bench, grab the ball, and bring it back to the owner.}\\
        \end{quote}
    In this example, the focus of the question (the dog) is visible in the video. To answer the question, one can use the information in the video (the ball on the bench) to deduce the series of actions the dog should take (running, jumping, grabbing, and returning the ball) in order to continue the game. 

    \item[SENTIMENT.] The category involves questions that focus on the sentiment, mood, attitude, or emotion displayed by one or more characters in the video (i.e., animated beings) toward other entities or events in the scene, throughout the entire video. A typical response to a sentiment question may describe a specific sentiment, attitude, or emotion, or it may reflect a neutral stance. This category represents a tool for evaluating model's ability to recognize and identify the emotional state or attitude of characters based on visual cues, reflecting their reaction or feelings toward the events and other entities in the video.\\
    Example: Suppose a video shows children who appear bored at a birthday party.
    
    \begin{quote}
            Italian:\\
    Q: \textit{Che atteggiamento hanno i bimbi?}\\
    A: \textit{Sono annoiati.}\\
        \end{quote}

    \begin{quote}
            English:\\
    Q: \textit{What is the attitude of the children?}\\
    A: \textit{They are bored.}\\
        \end{quote}
    In this example, the focus of the question (the children) is visible in the video. To answer the question, one can use the visual cues present in the video (expressions and behaviors of the children) to infer the sentiment (boredom) displayed by the characters.


    \item[SPATIAL.] Such a category involves questions related to the spatial relationships between entities, objects, or events depicted in the video. It aims at assessing the model’s ability to infer both stable and time-dependent spatial relationships, as well as the ability to determine relative positioning in space and to rely on grounding competencies.

\begin{itemize}
    \item \underline{Total Spatial}: This question asks about the position of entities in space (including their relation to other entities) that remains constant throughout the entire video, disregarding any temporal variations or minimal movements of the entity at different moments in the video. A typical response to this type of question provides specific spatial information valid for the entire duration of the video.\\
    Example: Suppose a video shows a school lesson with a teacher and students in a classroom.

        \begin{quote}
            Italian:\\
    Q: \textit{Dov'è l'insegnante?}\\
    A: \textit{L'insegnante è in classe.}\\
        \end{quote}

        \begin{quote}
            English:\\
    Q: \textit{Where is the teacher?}\\
    A: \textit{The teacher is in the classroom.}\\
        \end{quote}
    In this example, the focus of the question (the teacher) is visible throughout the video. To answer the question, one can use visible information from the video (classroom, desk) to provide the entity’s spatial position (behind the desk) throughout the video’s duration.
    
        \item \underline{Partial Spatial}: This question asks about the position of entities in space, but in relation to the time and/or other events occurring in the scene. It may also request the position of one entity relative to another, with a temporal aspect taken into account. A typical response provides spatial information that is specific only to the requested time range in the video.\\
        Example: Suppose a video shows a school lesson with a teacher and students in a classroom.
    
        \begin{quote}
           Italian:\\
    Q: \textit{Dove si trova l'insegnante all'inizio del video?}\\
    A: \textit{All'inizio del video, l'insegnante è di fronte la cattedra.}\\
        \end{quote}
        
        \begin{quote}
            English:\\
    Q: \textit{Where is the teacher at the beginning of the video?}\\
    A: \textit{At the beginning of the video, the teacher is standing in front of the desk.}\\
        \end{quote}
    In this example, the focus of the question (the teacher) is visible in the video. To answer the question, one would use the visual information visible in the specific part of the video (classroom, desk) to provide the spatial position (in front of the desk) relative to the time frame requested (at the beginning of the video).
    \end{itemize}


    \item[TEMPORAL.] The category includes questions that focus on temporal information. This category studies the model's ability to infer temporal relationships, sequence of events, and durations from visual content in a coherent manner.
    \begin{itemize}
        \item \underline{Partial Temporal}: This question focuses on the temporal properties and relationships of events in the video. The questions may request any type of temporal information about the events or their temporal relationships, except for their duration. For example, asking when something happens or if something happens before or after another event. A typical response provides the event with the specific temporal information requested by the question.\\
        Example: Suppose a video shows a rock band concert.
    
        \begin{quote}
            Italian:\\
    Q: \textit{Che succede dopo che il chitarrista inizia a suonare?}\\
    A: \textit{Il cantante inizia a cantare.}\\
        \end{quote}

        \begin{quote}
            English:\\
    Q: \textit{What happens after the guitarist starts playing?}\\
    A: \textit{The singer starts singing.}\\
        \end{quote}
    In this example, the focus of the question (what happens after the guitarist starts playing) is visible at a specific moment in the video. To answer the question, one can use the visible information in that portion of the video (the singer starts singing) to provide the event (the singer starting to sing) as a response.
    
        \item \underline{Duration Temporal}: This question focuses on a specific property of events in the video: their duration. A typical response provides the specific temporal information required by the question regarding the event’s duration.\\
        Example: Suppose a video shows a room with a light on and a person switching it off.
    
         \begin{quote}
            Italian:\\
    Q: \textit{Per quanto tempo la luce rimane accesa?}\\
    A: \textit{Per circa 15 secondi.}\\
        \end{quote}

        \begin{quote}
            English:\\
    Q: \textit{How long was the light on?}\\
    A: \textit{For about 15 seconds.}\\
        \end{quote}
    In this example, the focus of the question (the light on) is visible in the video. To answer the question, one can use the temporal information visible in the video (the person switching the light off) to provide a duration (about 15 seconds) as response.
    \end{itemize} 


    \item[UNCERTAINTY.] This question refers to entities and events that are part of the situation represented in the video, but the scene does not provide enough information to give a precise answer. Therefore, uncertainty questions involve a certain degree of ambiguity in the response, which cannot be fully derived from the video content. The answer may refer to a range of values, state that a precise answer cannot be given, or mention that the answer is a guess and might not be correct. This category tests the model's ability to recognise and deal with situations in which the available information is insufficient or ambiguous, leading to a response that reflects the uncertainty of the scene and indirectly testing the hypothetical assertive behaviour of such models in answering.\\
    Example: Suppose a video shows a dog.

    \begin{quote}
            Italian:\\
    Q: \textit{Quanti anni ha il cane?}\\
    A: \textit{Difficile da dire. / Il cane è probabilmente giovane, ma non si può esserne certi.}\\
        \end{quote}
    
        \begin{quote}
            English:\\
    Q: \textit{How old is the dog?}\\
    A: \textit{It's hard to say. / The dog is probably young, but it's not certain.}\\
        \end{quote}
    In this example, the focus of the question (the dog) is visible in the video. However, if one tries to answer the question, only partial information about the dog’s age is available in the video. As a result, an uncertain answer (e.g., “It's difficult to tell”) is expected. 

\end{description}

\subsection{True and False Statement Generation}
\begin{figure*}[h]
    \small
  \includegraphics[width=\linewidth]{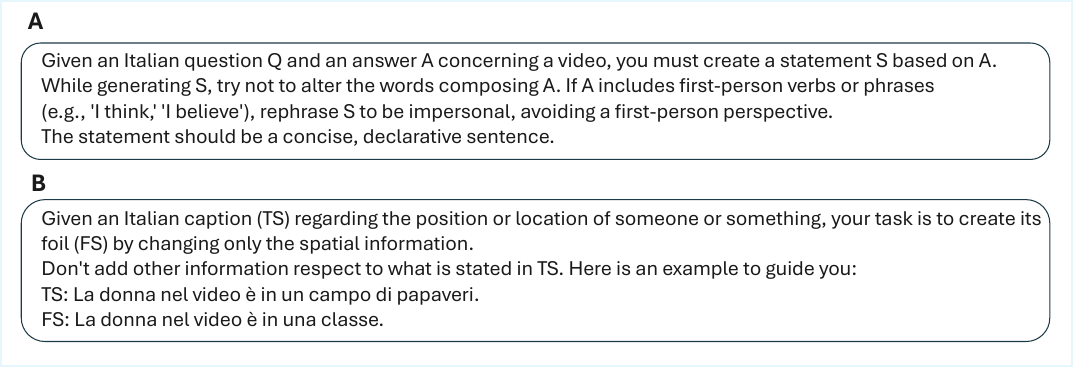}
  \caption{Prompts used for True (A) and False (B) Statements generation with GPT-4o. Prompt B is representative of the 12 different prompts used to generate False Statements, each tailored to a specific semantic category.}
  \label{fig:Prompt_StatementGen}
\end{figure*}

Figure \ref{fig:Prompt_StatementGen} illustrates the prompts used to generate True Statements (A) from the questions and the corresponding responses in the 8-answer pools and the False Statements (B) starting from the true ones.

 \begin{figure*}[h]
    \small
  \includegraphics[width=\linewidth]{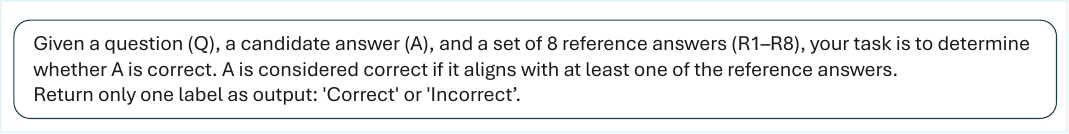}
  \caption{Prompt used for automatic evaluation of VLMs' answers in Task2 (i.e., for LLM-as-a-judge evaluation metric)}
  \label{fig:Prompt_llm-as-a-judge}
\end{figure*}

\section{Experiments}
\label{sec:appendixB}
This appendix section will contain additional details on our experimental settings, including a description of the VLMs used, as well as graphs and tables summarizing the results for Tasks 1 and 2 of MAIA. 
\\ In contrast to the initial experiments for creating and validating the synthetic data of the MAIA dataset, where we used \textit{OpenAI}'s \textit{GPT-4o} API, the experiments on MAIA were conducted using A$100$ GPUs ($40$GB). Overall, the total computational budget was on the order of $\sim$1,000 GPU hours.

\subsection{Models tested}
\label{sec:models_app}

\paragraph{Vision-Language Models}
We benchmarked six recent VLMs
. Both experiments and related quantitative evaluation have been done using \textit{lmms-eval} \citep{lmmevaluation}, a framework for the evaluation of multimodal models.
\begin{description}
[style=unboxed,leftmargin=0cm,noitemsep]
    \item[InternVL2.] \cite{internvl}: $8$B parameter transformer-based multimodal model employing advanced cross-attention; pre-trained on large-scale image-text and video datasets for diverse multimodal tasks and  instruction-tuned. It uses \textit{InternLM2.5} as open-sourced-$7$B parameter chat model. \textit{Hugging Face} model: OpenGVLab/InternVL2-8B.
    \item[InternVL3.] \cite{zhu2025internvl3exploringadvancedtraining}: $78$B parameter multimodal model trained with a native multimodal pre-training paradigm, integrating linguistic and visual capabilities from the start. It employs Variable Visual Position Encoding (V2PE) and advanced post-training strategies, achieving state-of-the-art performance among open-source VLMs. \textit{Hugging Face} model: OpenGVLab/InternVL3-78B.
    \item[LLaVA-NeXT-Video.] \cite{llavanextvideo}: $7$B parameter model built on the LLaVA framework, optimized for video understanding with mechanisms to capture temporal dynamics; fine-tuned on video instruction data. Base LLM: \textit{Vicuna}-$7$B (v$1.5$). \textit{Hugging Face} model: llava-hf/LLaVA-NeXT-Video-7B-hf.
    \item[LLaVa-OneVision.] \cite{llavaonevisioneasyvisualtask}: $7$B parameter model that builds on the LLaVA framework with a Qwen2 LLM backbone to serve as a general-purpose vision-language assistant; pre-trained on extensive multimodal data to deliver robust cross-modal reasoning. \textit{Hugging Face} model: lmms-lab/llava-onevision-qwen2-7b-ov.
    \item[Qwen2.5-VL.] \cite{qwen2.5technicalreport}: $7$B and $72$B parameter VLMs of the Qwen family using the \textit{Qwen}$2.5$ LLM decoder; key enhancements are related to grounding, working with longer videos and capturing events. It was pre-trained on comprehensive visual and textual datasets and fine-tuned for detailed, context-aware responses. \textit{Hugging Face} model: Qwen/Qwen2.5-VL-7B-Instruct and Qwen/Qwen2.5-VL-72B-Instruct.
    \item[Unimodal models.] As described in Section \ref{sec:exp_sett}
    we used five open-weight LLMs which have shown good performance on a variety of  tasks on Italian \cite{magnini2025evalitallmbenchmarkinglargelanguage}. For conducting these experiments, we used \textit{Minicons} library \cite{misra2022minicons}, a high-level wrapper around \textit{Hugging Face} for investigating predictive behavior of transformer models. Specifically, probabilities were computed adding a normalization parameter to take into account the different length of sentences in terms of tokens. Models used in the \textit{Hugging face} Hub are: \textbf{Llama-$3.1$} ($8$B-Instruct), \textbf{
LLaMAntino-$2$} ($7$B), \textbf{
LLaMAntino-$3$-ANITA} ($8$B-Instruct), \textbf{
Gemma} ($7$B) and \textbf{Qwen$2.5$} ($7$B-Instruct).
\end{description}

\subsection{Tasks Details}
\label{task_details}
\begin{figure*}[h]

    \centering
  \includegraphics[width=\linewidth]{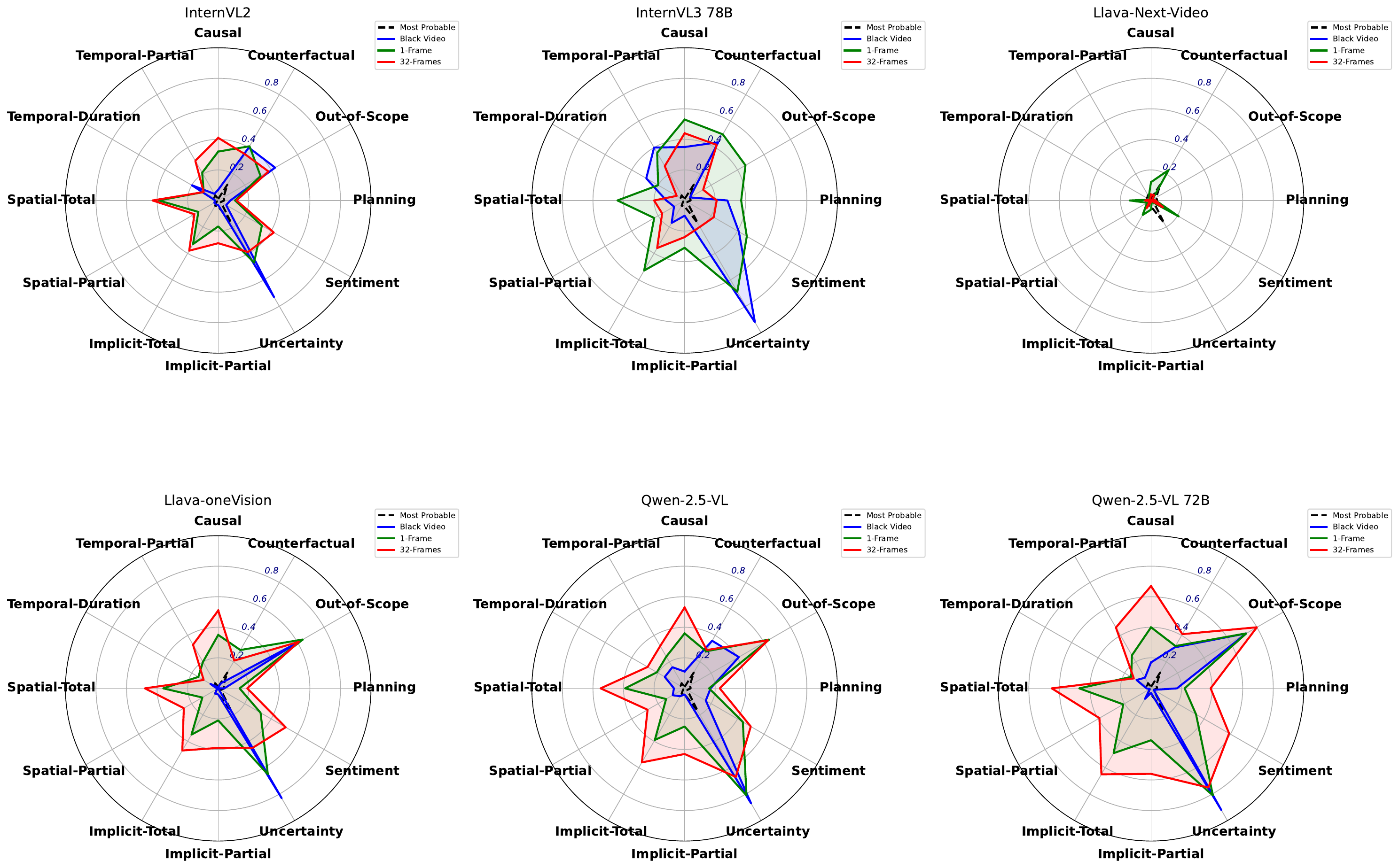}
  \caption{Task 1 pool-based}
  \label{fig:spider_task1_cons}
\end{figure*}

Table \ref{tab:task1_res_app} provide details with respect to the results obtained in Task 1 without considering any form of aggregation into pools (i.e. single-accuracy).
Figures \ref{fig:spider_task1}, \ref{fig:spider_task1_cons}, and \ref{fig:spider_task2} represent the fingerprint of models through MAIA's reasoning categories. Specifically, Figure \ref{fig:spider_task1} reports results for Task 1, Figure \ref{fig:spider_task1_cons} for Task 1 when models make 8/8 correct choices within the 8 TS-FS pairs that make up the pools, and Figure \ref{fig:spider_task2} for Task 2.\\
As regards the generation task, we combined similarity-based metrics with an LLM-as-a-judge approach, the latter being more suitable for handling open-ended responses. Using 8 reference answers for evaluating the generation correctness of VLMs allowed us to prioritize semantic alignment over surface similarity, following \cite{lee-etal-2024-prometheus} and \cite{vqa_evaluation_manos}. 
\textit{GPT-4o} was adopted as evaluation model, and its judgments showed high agreement with human annotators (Fleiss’ Kappa: 0.82). Figure \ref{fig:Prompt_llm-as-a-judge} shows the prompt used for this evaluation.

\begin{table*}[]
\resizebox{\textwidth}{!}{%
\begin{tabular}{lc|c|c|c|c|c|c|c|cc|cc|cc}
\toprule
 & \textbf{Models} & \textbf{Avg.} & \textbf{Causal} & \textbf{Counterfactual}  & \textbf{Out-of-Scope} & \textbf{Planning} & \textbf{Sentiment}&  \textbf{Uncertainty} & \multicolumn{2}{c|}{\textbf{Implicit}} & \multicolumn{2}{c|}{\textbf{Spatial}} & \multicolumn{2}{c}{\textbf{Temporal}} \\
 &  &  &  &   &  &  &  & & \textit{Partial} & \textit{Total} & \textit{Partial} & \textit{Total} & \textit{Duration} & \textit{Partial} \\ 
 \midrule

\textbf{Unimodal} &  {} & 0.56 & 0.45 & 0.73   & 0.55 & 0.68 & 0.53 & 0.79 & 0.50 & 0.51 & 0.48 & 0.51 & 0.53 & 0.48 \\ 
\midrule

\multirow{6}{*}{\textbf{Black video}} 
& \textit{InternVL2} 8B & 0.68 & 0.64 & \underline{\textbf{0.88}}   & 0.89 & 0.69 & 0.61& 0.96 & 0.52 & 0.59 & 0.54 & 0.55 & 0.67 & 0.60 \\
  & \textit{InternVL3} 78B & \textbf{0.77} & 0.81 & \underline{\textbf{0.88}} & 0.60  &  \textbf{0.80} & \textbf{0.84} & \underline{\textbf{0.99}} &  \textbf{0.64} & \textbf{0.71} & \textbf{0.68} & \textbf{0.71} & \textbf{0.78}  & \textbf{0.82} \\
 & \textit{Llava-Next-Video} 7B & 0.50 & 0.49 & 0.52   & 0.48 & 0.51 & 0.52 & 0.45 & 0.50 & 0.51 & 0.51 & 0.49 & 0.51 & 0.48 \\
 & \textit{Llava-oneVision} 7B & 0.59 & 0.54 & 0.48   & 0.92 & 0.61 & 0.38 & 0.97 & 0.52 & 0.53 & 0.50 & 0.52 & 0.60 & 0.51 \\
  & \textit{Qwen-2.5-VL} 7B & 0.73 & 0.68 & 0.86   & 0.87 & 0.75 & 0.72 & 0.98 & 0.56 & 0.62 & 0.61 & 0.65 & 0.73 & 0.70 \\
  & \textit{Qwen-2.5-VL} 72B & 0.69  & \textbf{0.71} & 0.78 & \underline{\textbf{0.95}}  &  0.75 & 0.57 & \underline{\textbf{0.99}} & 0.53  & 0.62 & 0.54 & 0.57 & 0.67  & 0.63 \\
\midrule

\multirow{5}{*}{\textbf{1-Frame}} & \textit{InternVL2} 8B & 0.75 & 0.80 & 0.87   & 0.69 & 0.71 & 0.77 & 0.89 & 0.65 & 0.80 & 0.65 & 0.82 & 0.67 & 0.69 \\
 & \textit{InternVL3} 78B & \textbf{0.81} & \textbf{0.86} & \underline{\textbf{0.88}}  & 0.81  & \textbf{0.83}  & \textbf{0.84} & 0.93 & 0.74 & 0.87   & \textbf{0.68} & 0.84 & \textbf{0.70} & \textbf{0.77}   \\
& \textit{Llava-Next-Video} 7B & 0.61 & 0.68 & 0.76   & 0.49 & 0.63 & 0.75 & 0.40 & 0.59 & 0.65 & 0.59 & 0.70 & 0.56 & 0.53 \\
& \textit{Llava-oneVision} 7B & 0.76 & 0.79 & 0.78   & 0.89 & 0.74 & 0.76 & 0.91 & 0.68 & 0.81 & 0.64 & 0.81 & \textbf{0.70} & 0.67 \\
& \textit{Qwen-2.5-VL} 7B & 0.79 & 0.81 & 0.81   & 0.91 & 0.75 & 0.82& 0.70 & \textbf{0.82}& \underline{\textbf{0.96}} & 0.67 & 0.82 & 0.69 & 0.73 \\ 
  & \textit{Qwen-2.5-VL} 72B & 0.79  & 0.81 & 0.81 & \textbf{0.92}  & 0.79  & 0.74 & \textbf{0.97} & 0.73 & 0.84 & 0.67  & \textbf{0.85} & \textbf{0.70} & 0.70   \\

 \midrule
 \midrule

\multirow{5}{*}{\textbf{32-Frames}} &  \textit{InternVL2} 8B & {0.79} & 0.83 &  0.85   &  {0.77} &  {0.75} &  {0.84}&  {0.84} &  {0.75} &  {0.83} &  {0.69} &  {0.86} &  {0.66} &  {0.76} \\
 & \textit{InternVL3} 78B & 0.64  & 0.81  & \textbf{0.87} & 0.55  & 0.71  & 0.59 & 0.62 & 0.61   &  0.76 & 0.67 & 0.71 & \underline{\textbf{0.79}}  & 0.82  \\
 &  \textit{Llava-Next-Video} 7B &  {0.52} &  0.56 &  {0.57}   &  {0.42} &  {0.57} &  {0.61} &  {0.32}&  {0.52} &  {0.59}&  {0.52} &  {0.56} &  {0.51} &  {0.48} \\
 &  \textit{Llava-oneVision} 7B &  {0.81} &  0.87 &  {0.78}   &  {0.88} &  {0.76} &  0.88 &  {0.85} &  {0.80} &  {0.85} &  {0.71} &  {0.87} &  {0.65} &  {0.80} \\
 &  \textit{Qwen-2.5-VL} 7B &  0.84 &  0.89 &  {0.80}   &  0.89 &  0.78 &  {0.86}&  0.92 &  0.82 &  0.88&  \underline{\textbf{0.83}} &  0.90 &  0.75 &  0.81 \\ 
  & \textit{Qwen-2.5-VL} 72B & \underline{\textbf{0.88}}  & \underline{\textbf{0.93}} & {0.86} & \underline{\textbf{0.95}}  & \underline{\textbf{0.85}}  & \underline{\textbf{0.89}} & \textbf{0.95} & \underline{\textbf{0.88}}  & \textbf{0.92}  & 0.80 & \underline{\textbf{0.92}} & 0.69  & \underline{\textbf{0.84}} \\
 \bottomrule
\end{tabular}%
}
\caption{\label{tab:task1_res_app}
    Visual statement verification (Task $1$): accuracy of correct choices across reasoning categories (without aggregation).
  }
\end{table*}

\begin{figure*}[h]

    \centering
  \includegraphics[width=\linewidth]{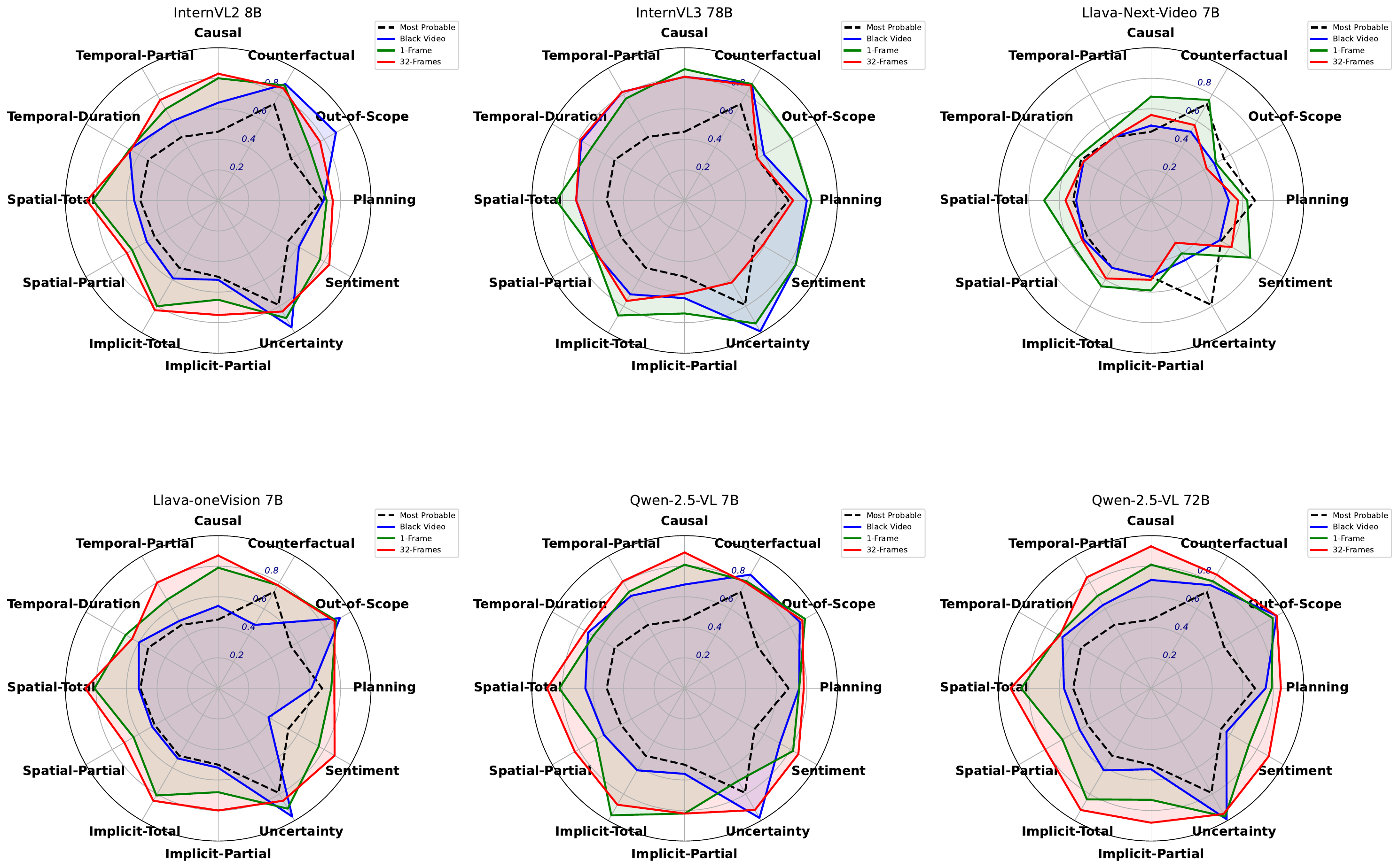}
  \caption{Task 1}
  \label{fig:spider_task1}
\end{figure*}

\begin{figure*}[h]

    \centering
  \includegraphics[width=\linewidth]{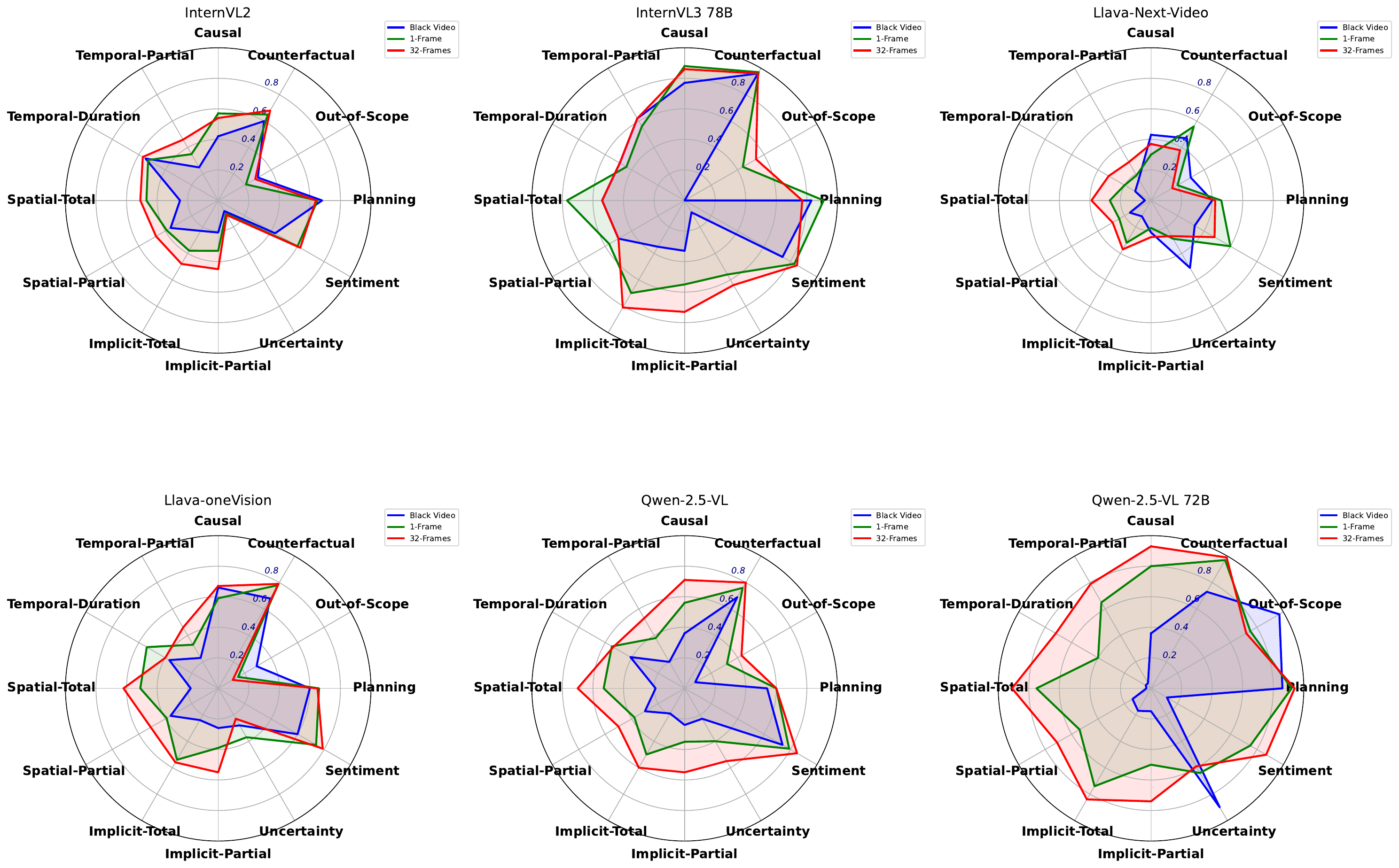}
  \caption{Task 2}
  \label{fig:spider_task2}
\end{figure*}

\subsection{Case study: Multimodal Hallucinations}
\label{sec:app_mhallucinations}

As part of a preliminary error analysis focusing on the \textsc{Out-of-Scope category}, we observed that a significant portion of the errors made by the model could be attributed to multimodal hallucinations. Particularly interesting was the discovery of counterintuitive clashes between the two tasks in our benchmark. In several instances, the model successfully solved Task 1 (i.e., True Statement Selection), in some cases achieving also full consistency within the pools (e.g., 8/8 correct selections), yet failed Task 2 (i.e., open-ended NLG), generating hallucinated content. For example, given a video in which a young girl is frightened by an insect in her home and seeks her mother's help to remove it, one of our best-performing model (i.e., \textit{Qwen2.5VL}) consistently selected the correct true statement in all eight pairs\footnote{True Statements are the first in each pair}:
\begin{enumerate}

    \item \begin{itemize}
        \item In the scene there is no dog at the door
        \item In the scene there is a dog at the door
    \end{itemize}
    
    \item \begin{itemize}
        \item In the movie there are no animals at the door
        \item In the movie there are animals at the door
    \end{itemize}

    \item \begin{itemize}
        \item In the movie there is no doggie at the doorway
        \item In the movie there is a doggie at the doorway
    \end{itemize}

    \item \begin{itemize}
        \item  No pets are seen in front of the door
        \item Some Pets are seen in front of the door
    \end{itemize}

    \item \begin{itemize}
        \item  There are no dogs at the door in the video
        \item There are dogs at the door in the video
    \end{itemize}

    \item \begin{itemize}
        \item  No dog appears in the entrance area
        \item A dog appears in the entrance area
    \end{itemize}

    \item \begin{itemize}
        \item  No pets are visible in the video near the front door of the house
        \item A pet is visible in the video near the front door of the house
    \end{itemize}

    \item \begin{itemize}
        \item  In the video clip there is no dog at the door
        \item In the video clip there is a dog at the door
    \end{itemize}

\end{enumerate}

However, when prompted in Task 2 with the question \emph{What color is the dog at the door?}, the model hallucinates by answering \emph{The dog at the door is black}, despite the fact that no dog is present in the video. This case highlights a curious but also dangerous misalignment between the model’s apparent ability to correctly perform discriminative reasoning in a multiple-choice setting and its failure to accurately generate grounded content. It also emphasizes a lack of robustness in the model’s competencies. These findings underscore the need for a more in-depth error analysis, as the overall results (see Tables 2 and 3) suggest that, for specific reasoning categories, performance in Task 1 may not reliably predict success in Task 2, with notable performance unbalances often occurring (e.g., \textsc{uncertainty} reasoning category).

\end{document}